\documentclass[notitlepage,12pt]{jedm}
\usepackage[table]{xcolor}
\pdfoutput=1
\usepackage{url}
\usepackage{makecell}
\usepackage{hyperref}
\usepackage{graphicx}
\usepackage[title]{appendix}
\usepackage{longtable}
\usepackage{placeins}
\usepackage{longtable}
\usepackage{changepage}
\usepackage{array}
\usepackage{enumitem}

\newcommand\MyBox[2]{
  \fbox{\lower0.75cm
    \vbox to 1.7cm{\vfil
      \hbox to 1.7cm{\hfil\parbox{1.4cm}{#1\\#2}\hfil}
      \vfil}%
  }%
}

\usepackage{titlesec}

\usepackage{booktabs}
\usepackage{subcaption}
\usepackage{float}
\usepackage{multirow}

\usepackage{amsmath}
\usepackage[noabbrev]{cleveref}
\hypersetup{
  colorlinks   = true, 
  urlcolor     = blue, 
  linkcolor    = blue, 
  citecolor   = blue 
}

\begin{document}

\title{Extending the Hint Factory for the assistance dilemma: 
A novel, data-driven HelpNeed Predictor for proactive problem-solving help}
\date{} 

\author{
{\large Mehak Maniktala}
\thanks{This reseach was supported by the NSF grants \#1726550 and \#1651909.}
\\North Carolina State University\\mmanikt@ncsu.edu 
\and  
{\large Christa Cody}\\North Carolina State University\\cncody@ncsu.edu
\and 
{\large Amy Isvik}\\North Carolina State University\\aaisvik@ncsu.edu   
\and 
{\large Nicholas Lytle}\\North Carolina State University\\nalytle@ncsu.edu  
\and 
{\large Min Chi}\\North Carolina State University\\mchi@ncsu.edu 
\and 
{\large Tiffany Barnes}\\North Carolina State University\\tmbarnes@ncsu.edu 
}

\maketitle

\begin{abstract}
Determining \emph{when} and \emph{whether} to provide personalized support is a well-known challenge called the assistance dilemma. A core problem in solving the assistance dilemma is the need to discover \emph{when} students are unproductive so that the tutor can intervene. Such a task is particularly challenging for open-ended domains, even those that are well-structured with defined principles and goals. In this paper, we present a set of data-driven methods to classify, predict, and prevent unproductive problem-solving steps in the well-structured open-ended domain of logic. This approach leverages and extends the Hint Factory, a set of methods that leverages prior student solution attempts to build data-driven intelligent tutors. We present a \textbf{\emph{HelpNeed}} classification, that uses prior student data to determine \emph{when} students are likely to be unproductive and need help learning optimal problem-solving strategies. We present a controlled study to determine the impact of an Adaptive pedagogical policy that provides \emph{proactive} hints at the start of each step based on the outcomes of our HelpNeed predictor:  productive vs. unproductive. Our results show that the students in the Adaptive condition exhibited better training behaviors, with lower help avoidance, and higher help appropriateness (a higher chance of receiving help when it was likely to be needed), as measured using the HelpNeed classifier, when compared to the Control. Furthermore, the results show that the students who received Adaptive hints based on HelpNeed predictions during training significantly outperform their Control peers on the posttest, with the former producing shorter, more optimal solutions in less time. We conclude with suggestions on how these HelpNeed methods could be applied in other well-structured open-ended domains.  \\  

{\parindent0pt
\textbf{Keywords:} assistance dilemma, problem solving, unproductivity, efficiency, unsolicited help, adaptive support, predicting help need, data driven tutoring, propositional logic, help appropriateness
}
\end{abstract}

\section{Introduction}
Intelligent Tutoring Systems (ITSs) provide individuals with adaptive feedback and hints, improving learning \cite{vanlehn2011relative}. Studies suggest that hints, when provided appropriately, can augment students’ learning experience \cite{bunt2004scaffolding,puustinen1998help} and improve their performance \cite{bartholome2006matters}. However, researchers often find that students display poor help-seeking behavior \cite{aleven2006toward,price2017hint}; some abuse hints to expedite problem completion, and some avoid seeking help when they are in need \cite{aleven2000limitations,price2017factors}. To ensure learning despite student help avoidance, several ITSs provide unsolicited assistance \cite{arroyo2001analyzing,murray2006comparison,kardan2015providing}. However, determining  \emph{when} and \emph{whether} to provide proactive assistance, i.e., unsolicited help in anticipation of future struggle, is particularly challenging in open-ended domains where there are many possible correct solutions.  More generally, it belongs to a well-recognized challenge in the domain of ITSs, called the assistance dilemma. 

The assistance dilemma is a trade-off between giving and withholding information to achieve optimal learning \cite{koedinger2007exploring}. A core problem of the assistance dilemma is the need to discover \emph{when} and \emph{whether} students are unproductive so that the tutor can intervene. Several studies have explored ways to determine the timing of assistance, as well as scaffolding in open-ended domains to improve learning \cite{fossati2015data,ueno2017irt,borek2009much,kardan2015providing}, and prevent student failure in exams \cite{merceron2005educational}. While some researchers have explored the generalizability of such approaches \cite{fratamico2017applying,bunt2003probabilistic}, determining when to provide assistance is still a challenging task for most open-ended domains, particularly because of differences in domains and learners \cite{klahr2009every,mclaren2014web}.

Open-ended domains can be ill-structured, where problems do not have a clear goal, set of operations, end states or constraints; or they can be well-structured, where problems have a clear goal, end states, or constraints. In this study, we seek to determine \textit{when} to provide unsolicited hints in a logic tutor to solve the assistance dilemma. While logic problems are well-structured in that they contain all the information needed to solve the problem and there are well-defined algorithms that students can use to solve them, they are open-ended in the sense that they have many possible solutions that can all be correct. Our approach builds upon the Hint Factory, a method for hint generation \cite{barnes2011using}. We extend the Hint Factory algorithm to define step-level productivity, and define \textit{HelpNeed}, to classify steps as unproductive or needing help. Based on this HelpNeed definition, we explore several machine learning methods to build a HelpNeed predictor and design an Adaptive pedagogical policy to proactively provide hints when the next step is predicted to need help. We conduct a controlled study to investigate the impact of Adaptive hints on student posttest performance, as well as the frequency of possible appropriate and inappropriate help during training, when compared to students in the Control group with access to on-demand hints but no Adaptive hints.

Our results showed that  students in the Adaptive condition have significantly higher posttest performance than those in the Control condition, as measured by solution optimality and time and the former also exhibit better hint usage behaviors while training on the tutor in that they had fewer HelpNeed steps, a lower possible help avoidance, and higher possible help appropriateness (a higher chance of receiving help when it was likely to be needed) when compared to the Control. Our main contributions can be summarized as:

\begin{itemize}
 \item This work presents a novel, data-driven HelpNeed classification model to determine unproductivity for steps, and a predictor for when unproductive steps might occur in well-structured, open-ended, multi-step problem-solving domains. 
 \item We develop a new Adaptive pedagogical hint policy that uses a HelpNeed predictor to provide proactive hints when they are likely to be needed, and investigate its impact on performance and appropriate help in a controlled study. 
\end{itemize}

In the remainder of the paper, we first discuss the related work that establishes the need for new ways to solve the assistance dilemma in well-structured open-ended problem solving. Next, we introduce the Deep Thought logic tutor, the context for this work. Then, we present our novel extensions to prior work on data-driven intelligence for tutoring to build HelpNeed classification models. We then compare machine learning methods to derive a HelpNeed predictor tuned to proactively identify steps where students are likely to need help. We present an Adaptive hint policy using the HelpNeed predictor, and describe our controlled study and its impact on posttest performance and whether it helped solve the assistance dilemma during training. 

\section{Related Work}
\subsection{Help Seeking}
Aleven et al. \cite{aleven2006toward} 
defined non-optimal help-seeking behaviors including \textit{help avoidance}, where students can benefit from seeking help but choose not to, and \textit{help abuse} where students excessively use help in situations where they could solve problems without assistance. 
While some studies aim to solve the problem of help avoidance by regulating students’ help-seeking behavior \cite{roll2011improving,aleven2006toward}, some employ unsolicited hints as we propose here \cite{arroyo2001analyzing,murray2006comparison,rus2017study}. Arroyo et al. observed that unsolicited hints could improve learning gains for students with low prior knowledge 
\cite{arroyo2001analyzing}. Murray et al. found that proactive help reduced frustration and saved students' time when they were struggling \cite{murray2006comparison}.

\subsection{Assistance Dilemma}
The \textit{assistance dilemma} has been a well-known challenge in providing unsolicited hints. Providing assistance may reduce frustration and save students’ time, but may lead to shallow learning or a lack of motivation to learn by oneself. On the other hand, withholding information can encourage students to learn by themselves, but may lead to frustration and wasted time \cite{koedinger2007exploring,mclaren2014web}. Researchers have investigated data-driven approaches to address this dilemma and create adaptive assistance both for \textit{on-demand} hints provided on request \cite{anohina2007advances,price2017factors,wood1999help} and \textit{proactive} unsolicited hints \cite{kardan2015providing,ueno2017irt}. Koedinger and Aleven \cite{koedinger2007exploring} worked towards addressing this dilemma in a cognitive tutor by initially withholding information about problem solutions and steps, and then interactively adding information, only as needed, through yes/no feedback, explanatory hints, and dynamic problem selection. Several studies have investigated approaches to resolve the assistance dilemma by modeling student behavior \cite{murray2004looking,ueno2017irt,conati2002using,kock2010towards}. One group created a tutoring policy within the DT Tutor that applied decision theory to make its choice about whether to give a hint \cite{murray2004looking}. In a study conducted by Ueno et al. \cite{ueno2017irt} in a programming tutor, they modeled student performance using Item Response Theory to determine optimal hint scaffolding for providing assistance proactively upon mistakes. The authors discovered that hint scaffolding with predicted probabilities for the learners' success to be 0.5 provides the best learning performance. We recently added unsolicited hints to our tutor, improving post-test performance for students with low prior knowledge \cite{maniktala2020leveraging}. However, the posttest performance of students with high prior knowledge was negatively, but not significantly, correlated to how often they used the unsolicited hints, suggesting the potential for negative effects from receiving too many hints. Therefore, in this work we seek to develop adaptive support that can determine \textit{when} to proactively provide hints based on students’ aptitude or performance, as in other ITSs \cite{bunt2003probabilistic,kardan2015providing}. 

The assistance dilemma is particularly challenging to address in open-ended domains that deal with ill-structured problems \cite{mclaren2014web,borek2009much} as well as well-structured problems \cite{ueno2017irt}. McLaren et al. in \cite{mclaren2008and} conducted a study to explore the assistance dilemma in an ill-structured chemistry tutor with inquiry learning using three levels of problem-level assistance: high (worked examples), mid-level (less assistance than worked examples), and low (untutored problem solving). They found mid-level assistance to lead to better learning than either low- or high-level assistance. In another study, Borek et al. in \cite{borek2009much} investigated the optimal amount of step-level assistance in a discovery-oriented virtual chemistry laboratory (VLAB) and found that students learned the conceptual tasks better with a mid-level assistance approach of hints and feedback. Further, their results suggested that assistance should be given only when students are far off track. 

Kardan and Conati used clustering and mining association rules to model when to provide unsolicited adaptive support (hint timing, content, and interface) in an ill-structured exploratory interactive simulation environment for teaching constraint satisfaction problems \cite{kardan2015providing}. While the students who received adaptive support did not have improved task performance, they learned significantly more than those who did not receive the adaptive support. Fratamico et al. in \cite{fratamico2017applying} applied Kardan and Conati’s 2015 framework to an electronic circuits simulator and found that it successfully classified students into groups of high and low learners. Mostafavi, et al. similarly applied machine learning to form data-driven proficiency profiles used in problem selection \cite{mostafavi2015data}, and showed that it reduced the time taken in a previous version of the Deep Thought logic tutor. Conati et al. in \cite{conati2002using} presented several evaluations of a Bayesian Network used to model student behavior in a well-structured physics tutor that, in part, is used to determine when unsolicited mini-lessons should be provided to students. They also described how these modeling tasks involve a high level of uncertainty when students are allowed to follow various lines of reasoning and are not required to explicitly show their reasoning. As mentioned above, a study conducted by Ueno et al. \cite{ueno2017irt} in a well-structured programming tutor used Item Response Theory to determine optimal scaffolding for unsolicited hints and discovered that providing scaffolding associated with 50\% probability for students’ success led to the best student outcomes.

\subsection{Unproductive Behavior}
Several studies have used the term “unproductive” to refer to undesirable behavior during training \cite{kai2018decision,botelho2019developing,park2018predicting}. For example, Beck and Gong \cite{beck2013wheel} define unproductive persistence or “wheel-spinning” based on whether or not a student achieved mastery (three correct problems) in ten problem attempts. Their definition of unproductivity has been used in recent studies to predict when an intervention can help students by distinguishing between productive and unproductive behavior using decision trees \cite{kai2018decision} and recurrent neural networks \cite{botelho2019developing}. However, this definition of problem-level and problem-completeness based productivity is not suitable for our objective of guiding students toward efficient problem-solving strategies at a more granular step-level, specifically in domains where problems can have several solution paths. 
 
In a study on an open-ended and ill-structured inquiry-learning program, McLaren et al. defined unproductive events as student actions that are unlikely to advance understanding \cite{mclaren2014web}. Similarly, we identify unproductive problem-solving steps that are not likely to advance student’s problem-solving strategies. However, our definition is 
based on time and solution length/optimality, rather than domain-specific productivity models.

In many multi-step open-ended but well-structured problem-solving domains, shorter solutions are considered to be more optimal, and solving problems in less time reflects both learning and fluency \cite{mayer1992thinking,smith2012toward,yacef2005logic,cen2007over}. We use these basic assumptions about time and solution length to design a data-driven approach to model productivity on problem-solving steps, extending the Hint Factory, a data-driven approach for hint generation. This 
approach uses prior students' 
data to 
assign scores to problem-solving states 
\cite{stamper2008hint,barnes2011using}. A core insight of this paper is that we can similarly use student data to score productive problem-solving steps without the need to model the domain. 

Some studies have observed that students game educational systems to elicit unsolicited hints to complete problems faster \cite{baker2009educational,d2006adapting,murray2005effects}. Therefore, we limit the amount of information provided in our unsolicited hints, presenting them as partially-worked steps. 

\section{Tutor Context}
\label{sec:dt}

The research in this paper is conducted using Deep Thought, a data-driven intelligent logic tutor where students practice constructing formal propositional logic proofs in discrete math courses, with 200-350 students per semester, since 2008 \cite{stamper2008hint,barnes2010automatic,cody2017investigating,mostafavi2017evolution}. Each fall (f) or spring (s) semester, we use stratified sampling on a pretest to assign students to conditions. Historical data exist for 2017, 2018, and 2019. This study uses data from f17, s18, f18, and s19; the HelpNeed predictor uses f18, and s19; and the controlled study with the proactive hint policy was in f19. Finally, the results of the controlled study are analyzed comparing data from f18R, s19R, along with f19, where R represents a subset of the population that received \textit{randomly} administered frequent unsolicited hints\footnote{With a constraint that three unsolicited hints cannot be given in a row}\cite{maniktala2020leveraging}. Each section provides more details on the specific datasets used. 

\begin{figure*}
\centering
\includegraphics[width=0.90\columnwidth]{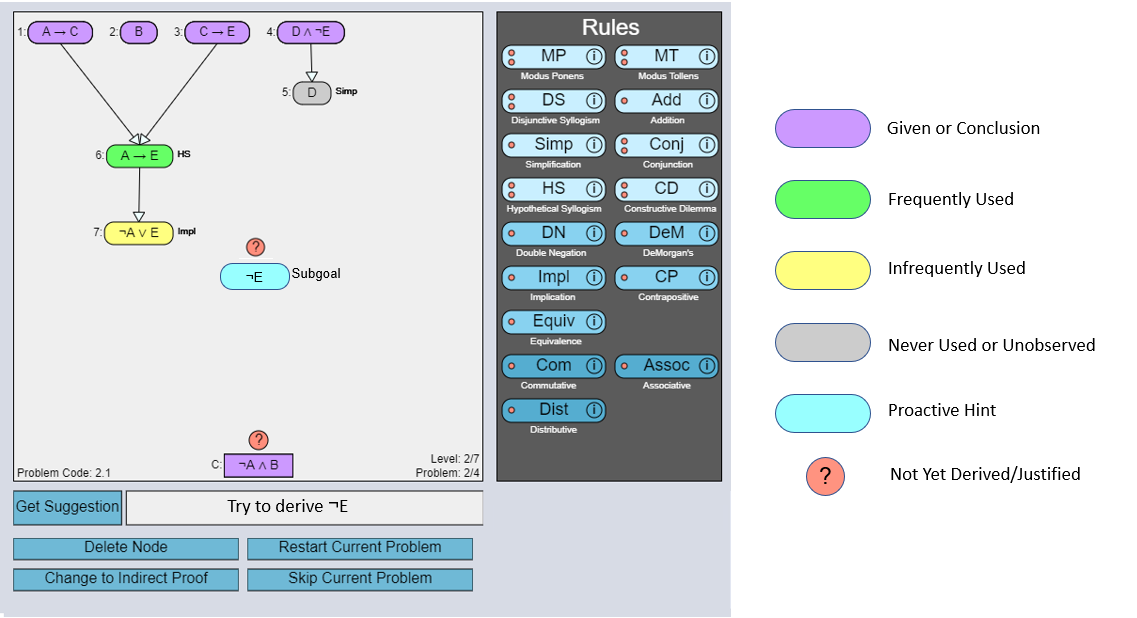}
\caption{Deep Thought interface: \textit{Workspace} (left) with the provided givens (top) and the conclusion (bottom), domain rules (middle), the \textit{`Get Suggestion’ hint button} and \textit{hint message box} (bottom-left). The color-coding is explained on the right (provided to students in an infobox).}
\label{fig:dt_ui}
\end{figure*}

Figure \ref{fig:dt_ui} shows the Deep Thought interface for a single problem, where students construct their proofs in the \textit{workspace}. Each proof problem starts with a set of given logic statements with a conclusion to derive. Each statement is represented by a \textit{node}, which students can click to select and apply a logic rule to derive a new node. Each node derivation consists of two parts: the \textit{justification} and the derived statement. The justification is a set of 1-2 existing nodes and the rule applied to them, and the derived statement is the result. The tutor checks each step's justification and derived statement to provide immediate feedback on the correctness of their logic rule applications. In case of rule application errors, students receive a notification of the mistake via popup,  but the attempted incorrect node is deleted from the workspace. When a statement is correctly justified and derived, a correct step is recorded and its node appears. The nodes are colored according to their \textit{use-frequency}\footnote[1]{A node is said to be \textit{used} if its removal from the solution would make the problem incomplete.} in historical student solutions. A frequently-used node is colored green, an infrequently-used node is colored yellow, and a node that has never been used or has not been observed in the historical data is colored gray (see Figure \ref{fig:dt_ui}). Hints are available during Deep Thought’s training section using the Hint Factory \cite{stamper2008hint}. Students can receive \textit{on-demand} hints below the workspace by clicking the “Get Suggestion” button, and they can also receive tutor-initiated \textit{proactive} hints that appear within the workspace (see Figure \ref{fig:dt_ui}).

A snapshot of the workspace at any given time is called the problem-solving \textit{state}, and transitions between states are called \textit{steps}. When a problem is complete, students have connected the given statements to the conclusion. 


\section{HelpNeed Classification and Predictor} 
\label{stage1}
In this section, we present our methods to determine when students need help learning efficient problem-solving strategies. 
We seek to identify unproductive steps where students need help, so that we can predict them, and provide proactive hints with the aim to prevent them.
To do this, we first define our HelpNeed classification based on prior literature on problem-solving, and later present our HelpNeed predictor to detect the need for help at the start of each step. 

\subsection{HelpNeed Classification}
We designed the HelpNeed classification based on the literature on what behaviors may demonstrate a detectable need for help during problem-solving. Based on the literature, learning is reflected in both correctness and/or time (duration) in problem solving \cite{corbett1994knowledge,kai2018decision,beck2013wheel}. However, since multi-step problem-solving involves several steps in a row, we cannot directly use correctness as a criterion. Therefore, we create new methods to detect efficiency, which is a proxy for eventual correctness and optimality in multi-step problem solving, as discussed in detail in section \ref{sec:efficiency}. 

Table \ref{tab:def_unp} shows our HelpNeed classification, which is based on a step’s duration and efficiency. Duration is the time taken by a student to carry out a step, described in detail below in \ref{sec:duration}. Efficiency is our unique extension of the Hint Factory, that uses prior data to assign numeric values that reflect a step’s quality and how well it promotes progress to a good solution, described in \ref{sec:efficiency}. 

The HelpNeed classification scheme is designed to identify steps that reflect suboptimal strategies, i.e., steps that unnecessarily increase the solution length and/or problem-solving time. 
We first define steps that do not demonstrate HelpNeed. For our first two categories, efficient steps, that lead to shorter, more optimal solutions, are considered productive irrespective of duration, with short times representing (1) \textit{Expert-like} behavior and longer times classified as (2) \textit{Strategic}. 

For our third category, we classify a single quick but inefficient step, i.e. a plausible guess, as an (3) \textit{Opportunistic} step.  Research suggests that we should provide students some opportunities to guess throughout the learning process in semi open-ended domains such as math \cite{capraro2012investigation,polya2004solve}. Polya  \cite{polya2004solve} describes guess-and-check to be a strategy when students apply \textit{plausible reasoning}, where the goal is to distinguish a more reasonable guess from a less reasonable one. In a beneficial use of guess-and-check, the student \textit{checks} whether they are moving closer to the goal and adjusts their strategy in the next step. Therefore, we classify a single quick but inefficient, Opportunistic step as one where help is \textit{not needed}. To transfer the idea of Opportunistic steps to other domains, it will be important to determine what combination of efficiency and duration constitutes a plausible guess, or Opportunistic step, where help is not needed, in contrast to the remaining categories below. 

Our fourth (4) \textit{Far Off} category is meant to represent one in a series of steps that demonstrate a lack of strategy, or being far off-track, and needing help. A study by Borek et al. in an ill-structured open-ended domain, suggests that students who are far off track in problem-solving can benefit from a hint \cite{borek2009much}. Likewise, in the well-structured, open-ended domain of linked lists, Fossati et al. intervened when students made steps that prevented problem completion \cite{fossati2015data}. Research suggests that repeated guessing when faced with difficulty can be a significant stumbling block in developing effective learning strategies, and feedback at this stage can help students \cite{kinnebrew2014analyzing}. 

Based on the meaning of Far Off in these studies, we have defined Far Off steps as those that demonstrate that the student is not reflecting on whether their consecutive steps are leading toward an efficient solution. In our tutor, evidence of being Far Off can be seen when a student quickly performs sequential inefficient steps. Therefore, we define Far Off steps as the second and later steps in a series of quick, inefficient steps. In a sequence of three quick inefficient steps, the first would be Opportunistic - with no need for help, and the second and third would be classified as Far Off - with help needed. In this tutor, where the mean student solution length is 10 steps, we determined that 2 steps are about 20\% of the problem, and this is long enough to intervene. The definition of Far Off steps could be adjusted for other domains to align with being far off track, using a combination of duration and efficiency, for example by setting a maximum time or number of inefficient steps that can elapse without an efficient step being taken. 

Next, researchers use the term ``wheel-spinning" or unproductive struggle to refer to the lack of mastery in a timely manner \cite{beck2013wheel,kai2018decision}. In our final category, a student who has spent significant time but has derived a step that is inefficient may be engaging in unproductive struggle, and we classify this as a (5) \textit{Futile} step where a student needs help.  In the next two sections, we describe the step duration and efficiency components of HelpNeed classification.

\begin{table}[!ht]
\centering
\caption{Defining HelpNeed using step Efficiency and Duration}
\begin{tabular}{|l|l|l|}
\hline
Classification                & Behavior      & Description                                                                                                                    \\ \hline
\multirow{3}{*}{No HelpNeed} & Expert-like   & A quick efficient step; demonstrating mastery                                                                                    \\ \cline{2-3} 
                              & Strategic     & \begin{tabular}[c]{@{}l@{}}A long efficient step; taking longer on an \\ expert-like step\end{tabular}                           \\ \cline{2-3} 
                              & Opportunistic    & \begin{tabular}[c]{@{}l@{}}A single, quick inefficient step \end{tabular}                        \\ \hline
\multirow{2}{*}{HelpNeed}    & Far Off & \begin{tabular}[c]{@{}l@{}} Maximum number of inefficient steps in a sequence\\ and/or multi-step duration for inefficient steps before \\intervention is desired\\ In our tutor: consecutive quick but inefficient steps \end{tabular}           \\ \cline{2-3} 
                              & Futile        & \begin{tabular}[c]{@{}l@{}}A long inefficient step; taking too long on \\ a step that does not help make progress\end{tabular} \\ \hline
\end{tabular}
\label{tab:def_unp}
\end{table}

\subsubsection{Step Duration}
\label{sec:duration}
A step duration is said to be \textit{Long} if it is carried out in a time greater than $75^{th}$ percentile of step time for that problem, and  \textit{Quick} otherwise. These values are computed using historical per-problem data averaged across solutions by students ($\mathit{N} = 437$), to account for the longer step times in more difficult problems. For example, a step is Long in a difficult problem of our logic tutor when the step time is greater than 5.48 min, whereas a step is Long in one easy problem of our tutor if the step time is greater than 2.95 min. 


\subsubsection{Step Efficiency}
\label{sec:efficiency}
We define \textit{step efficiency} as a data-driven measure of how much a student’s most recent step contributes to an efficient (short) solution. We explore four new step efficiency metrics defined using the combination of the \textit{quality} of each state (local or global), and the \textit{progress} made in a step (absolute or relative). We first present the Hint Factory, and then how we extended it to define step efficiency. 

\begin{figure*}[h!]
\centering
\caption{Three Solutions with Varying number of steps for a logic problem in Deep Thought}
\begin{subfigure}{.29\textwidth}
  \centering
  \includegraphics[width=\linewidth]{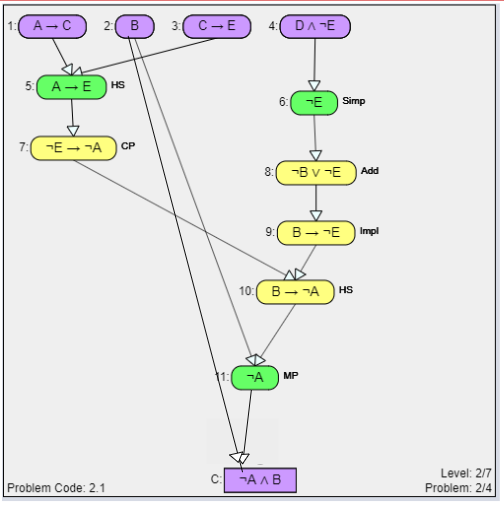}
  \caption{Trajectory $T_{long}$: Eight\\steps and all used nodes}
 \label{fig:t1}
\end{subfigure}
\begin{subfigure}{.29\textwidth}
  \centering
  \includegraphics[width=\linewidth]{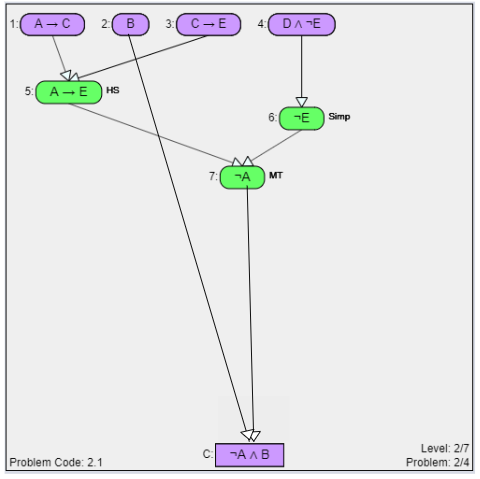}
  \caption{Trajectory $T_{short}$: Four\\steps and all used nodes}
  \label{fig:t2}
\end{subfigure}
\begin{subfigure}{.29\textwidth}
  \centering
  \includegraphics[width=\linewidth]{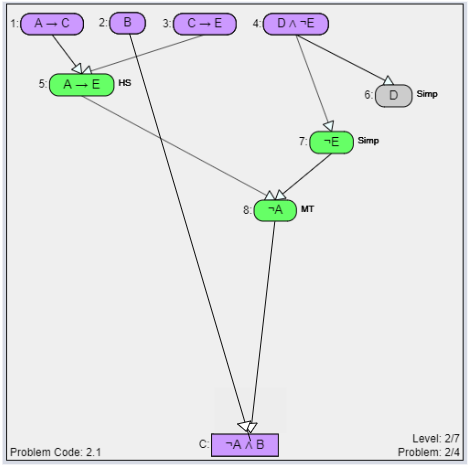}
  \caption{Trajectory $T_{medium}$: Five\\steps and four used nodes}
  \label{fig:t3}
\end{subfigure}

\label{fig:three_trajectories}

\vspace{2em}
\includegraphics[width=0.65\columnwidth]{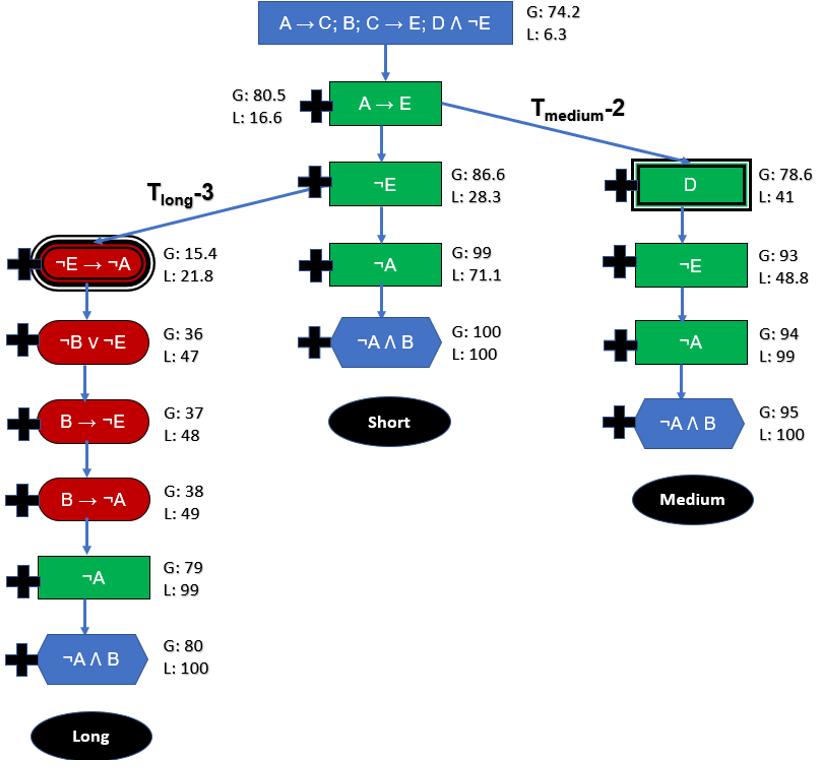}
\caption{Illustration for the concepts of State Quality and Productive steps in 3 trajectories $T_{short}$, $T_{medium}$, and $T_{long}$}
\label{fig:global_local}
\end{figure*}

\paragraph{The Hint Factory}\mbox{} \\
The Hint Factory is a method for generating hints in well-structured open-ended domains \cite{barnes2008pilot}. In this approach, historical student solutions are used to form Markov decision processes (MDPs) from interaction networks \cite{eagle2015interaction}, where vertices are observed student problem-solving states (snapshots of their on-going or completed proof), and edges are problem-solving steps, i.e, a transition between states.  The Hint Factory uses value iteration, a classic reinforcement learning technique, given in Equation \ref{eq:local_quality}, to assign an \textit{expected value} $V(s)$ to each state $s$, where $R(s)$ is the state’s  reward, $\gamma$ is the discount factor, and $P(s|s’)$ is the proportion of the observed solutions in state $s$ that lead to state $s’$ using the action $a$. An action in this equation is what causes a transition between states, therefore, action $a$ represents a problem-solving step, that can be carried out either by correct rule applications or by node deletions. For example, in Figure  \ref{fig:global_local}, the arrows represent actions (steps). More specifically, an action $a$ to derive node $A \rightarrow E$ from the top start state is a tuple consisting of the rule, the node list the rule applies to, and the resulting logic statement (HS, {$A \rightarrow C$, $C \rightarrow E$},$A \rightarrow E$). Note, while the Bellman Equation \ref{eq:local_quality} is used for stochastic environments in reinforcement learning, these actions (steps) are carried out deterministically in our tutor. The original Hint Factory used this equation not to account for the uncertainty in carrying out steps, but rather to take into account the probabilities of transition between each pair of states. This ensures that the value of each state is dependent on the probabilities of transition at each successive step. 
In the Hint Factory, a large reward is set for the problem-completion or \textit{goal states} (100), penalties for incorrect states (10), and a cost for taking each action (1) \cite{barnes2011using}. A non-zero cost on actions causes the MDP to penalize longer solutions.

\begin{equation}
  V(s) := R(s) + \gamma \max_{a} \sum\limits_{s'}P(s'|s)V(s')
              \label{eq:local_quality}
\end{equation}

\paragraph{State Quality - extending the Hint Factory}\mbox{} \\
In this section, we leverage the Hint Factory approach to generate two quality metrics that determine the expected values for each observed problem-solving state. The first metric of state quality is our prior work on the Hint Factory, which we label as \textit{local quality value} ($LQV$) with local rewards ($LR$). Local quality provides insights about how far a state is from the closest goal state, weighted by the probabilities of transitions, but it cannot provide information about whether the state is on an efficient path to a solution.

\begin{equation}
  GQV(s) := GR(s) + \gamma \sum\limits_{s'}P(s'|s)GQV(s')
              \label{eq:global_quality}
\end{equation}

\textbf{Global Quality}. We devised a novel, data-driven global quality value function, $GQV$ in Equation \ref{eq:global_quality}, to give higher values to states on efficient solution paths. Equation \ref{eq:global_quality} sums $GQV(s’)$ over all states $s’$ reachable from $s$, weighted by $P(s|s’)$, taking into account all future actions from a current state, rather than just the one with the max expected value. The global rewards $GR$ are identical to $LR$ for errors and actions, but are different for goals, giving shorter, more efficient solutions higher rewards. The global reward $GR(g)$ for each goal state $g$ on a problem is $GR(g) = 100 - p* \delta(g)$ where $\delta(g)$ is the difference between the solution length of $g$ and that of the shortest solution, and \textit{p} is a penalty for longer solutions. We set $p = (100-80)/ \delta_{median})$ where $\delta_{median}$ is the difference between the median and shortest solution lengths for each problem because median student solution lengths are assigned a global reward of 80. Our intention in setting this value of 80 is so that  the student's performance with a median solution length represents a low B grade. Note, the relationship between the action $a$ and follow-up state $s’$ is the same for both local and global state values, because performing the action $a$ when in state $s$ deterministically leads to the next state $s’$ based on our definition of actions. And, finally, the proof of convergence for the modified value iteration equation \ref{eq:global_quality} is given in appendix \ref{aconv}.

We now demonstrate the differences between local and global quality metrics using three solution trajectories (series of steps) of varying solution lengths: $T_{short}$, $T_{medium}$, and $T_{long}$ in Figures \ref{fig:three_trajectories} and \ref{fig:global_local}. $T_{short}$ is the shortest solution (four steps), with all nodes (derived problem-solving statements) used. Recall as defined in section \ref{sec:dt}, a node is said to be \textit{used} if it contributes towards deriving the conclusion of the problem. $T_{short}$ have four used steps, $T_{medium}$ has five steps with one unused node $D$; and $T_{long}$ has eight steps, and all nodes used.

We generated interaction networks to determine the quality values for each problem using our historical data for N = 796 students. Figure \ref{fig:global_local} shows the quality values for the three trajectories in Figure \ref{fig:three_trajectories}. The start state in Figure \ref{fig:global_local} consists of the four given statements (the topmost state). Arrows between states represent steps, i.e, a transition between states by rule applications. Non-start states are represented by a +(XYZ), where XYZ is the new statement derived in a step. The start state has a high global quality, but low local quality. The starting state’s global quality is high because all efficient paths contain it, but its local quality is low because it is probabilistically farther away from goals than any other state in the figure. The local quality for states that are only found in incomplete attempts is lower than that for the start state. The local quality of the goal states on all three trajectories is 100. The global quality value for the goal state in each solution trajectory differs, with 100 for the $T_{short}$ goal (since it’s the most efficient), 95 for $T_{medium}$, and 80 for $T_{long}$ goal states. 

From the start state to the goal in $T_{short}$, both local and global quality state values increase monotonically since it is the most efficient solution, and the chances of finishing the proof with an optimal solution increase with each step on this trajectory. Note that not all quality values increase over every trajectory. For example, step $T_{medium}-2$’s pre-state ($+ A \rightarrow E$) global quality is higher than that for its post-state ($+ D$) since the pre-state is probabilistically on a more efficient path, but the local quality increases from pre- to post-state. Step $T_{long}-3$’s pre-state ($+ \neg E$) has higher local and global quality values than its post-state ($+ \neg E \rightarrow \neg A$) since the post-state is farther from and less likely to reach $T_{long}$’s goal than the pre-state is to $T_{short}$’s closer goal. The global quality decreases for two reasons: (1) $T_{long}$ goal is on a less efficient path, and (2) global quality performs a weighted sum over all the subsequent, previously-observed states in the larger (unshown) interaction network, many of which lead to incomplete attempts.

These three example trajectories demonstrate the differences between local and global state quality metrics. The main strength of generating these quality values is the MDP approach which ensures that each state quality value is based not only on the distance from a solution, but also on the probability of transition at each of the successive steps. This allows us to rate steps in a more probabilistic manner than a simple comparison based on the distance from the most efficient expert solution.

\paragraph{Progress - Change in Quality}\mbox{} \\
Since state quality is a measure of relative ``goodness", we \textit{compare} the quality of the current state with that of the previous and start states to evaluate the efficiency of a problem-solving step. In this section, we define two measures for \textit{progress}: \textit{relative}, the change in state quality from the \textit{previous} problem-state, and \textit{absolute}, the change in state quality from the \textit{start} state.

\textit{Relative progress} is the difference between the quality values of the current and previous states. Relative progress with local quality identifies whether the previous or current state is probabilistically closer to the goal. When using the global quality values, the relative progress identifies which state is probabilistically closer and on a more efficient solution path.

Consider a valid, but long solution attempt. A relative progress measure reveals whether a student is progressing toward a solution in a step, but not whether their trajectory is efficient. Therefore, we define \textit{absolute progress} as the difference between the current and start states' quality, using either quality measure. Absolute progress using local quality reveals whether a student’s current state is probabilistically farther or closer from any goal states than when they began working on the problem. Global quality based absolute progress reveals the amount of efficient progress a student has made since the problem started. For example, if a student is always taking efficient steps, then the absolute progress will increase on every problem-solving step. 



\paragraph{Step Efficiency - Quality \& Progress}\mbox{} \\
We define four kinds of step efficiency measures based on quality \{Local, Global\} and progress \{Relative, Absolute\}. A step is considered \textit{efficient} if the progress of its post-state using either quality measure is a non-negative number, and \textit{inefficient} otherwise, as shown in Table \ref{tab:def_prod}.

\begin{table}[]
\centering
\caption{Step Efficiency formula based on state Quality and step Progress}
\label{tab:def_prod}
\begin{tabular}{|c|l|}
\hline
  Quality-Progress  & \multicolumn{1}{c|}{Efficiency Formula}               \\ \hline
Global-Absolute & ($GQV_{post-state}$ - $GQV_{start-state}$) $\geq$  0  \\ \hline
Global-Relative & ($GQV_{post-state}$ - $GQV_{pre-state}$) $\geq$ 0 \\ \hline
Local-Absolute  & ($LQV_{post-state}$ - $LQV_{start-state}$) $\geq$ 0   \\ \hline
Local-Relative  & ($LQV_{post-state}$ - $LQV_{pre-state}$) $\geq$ 0 \\ \hline
\end{tabular}

\end{table}

We now compare the four kinds of step efficiency using three solution trajectories ($T_{short}$, $T_{medium}$, $T_{long}$) shown in Figure \ref{fig:three_trajectories}. We checked our ratings of inefficient and efficient steps with an expert logic instructor. Steps that were rated by our expert as inefficient are displayed in red and others are in green in Figure \ref{fig:global_local}. According to the local quality and absolute progress (local-absolute) efficiency metric, all the steps are efficient because they eventually lead to a solution. However, this metric is not sensitive to variations in solution lengths. When we use local-relative efficiency, only the $T_{long}-3$ step is inefficient, as it is the only step where a post-state is probabilistically farther from a solution than the pre-state. Using the global-relative measure, steps $T_{medium}-2$ and $T_{long}-3$ are inefficient because they have a pre-state on a more efficient path to the solution than the post-state. The global-absolute metric is the only measure that labels the four expert-identified inefficient steps correctly\footnote[1]{Note that these four inefficient states also correspond to the four infrequently used (yellow) nodes in the student solution shown in Figure \ref{fig:t1}. However, some inefficient nodes have been observed to be frequently used, and some efficient nodes to be infrequently used in our tutor, suggesting that the use-frequency alone cannot determine step efficiency, as was done in our prior work \cite{stamper2009unsupervised}}. Note that each type of efficiency captures a different perspective on a step towards the solution. The global-absolute efficiency metric aligns the most with expert labels for the sample trajectories and our expert verified this alignment on a random sample of trajectories. Note that it would not be feasible to perform expert ratings on all student-derived steps -- even in small logic proofs we examined here, we have N = 72,560 unique states in the prior student data for 35 problems and 796 students.

\paragraph{Selecting a Step Efficiency metric}\mbox{} \\
\label{sec:select_cor}

\begin{table}
\centering
\caption{HelpNeed detected in historical data using each type of Step Efficiency and their Correlation with students’ posttest optimality (all correlations are significant with \textit{p} $<$ 0.01)}
\begin{tabular}{|c|c|c|}
\hline
\multirow{2}{*}{\begin{tabular}[c]{@{}c@{}}Step efficiency used to\\ define HelpNeed\end{tabular}} & \multirow{2}{*}{\begin{tabular}[c]{@{}c@{}}\% of HelpNeed steps \\ w/ this metric\end{tabular}} & \multirow{2}{*}{\textit{Corr}} \\
                                                                                                               &                                                                                                               &                                \\ \hline
Global-Absolute                                                                                                & 24.86                                                                                                         & -0.36                        \\ \hline
Global-Relative                                                                                                & 23.76                                                                                                         & -0.32                         \\ \hline
Local-Absolute                                                                                                 & 16.09                                                                                                         & -0.29                         \\ \hline
Local-Relative                                                                                                 & 18.14                                                                                                         & -0.31                         \\ \hline
\end{tabular}

\label{tab:corr}
\end{table}
To understand which one of the four step efficiency metrics is most indicative of how students' work in the tutor's training section affects their posttest solution optimality, we conducted a correlation test. Note, we evaluate students on an \textit{optimality} score based on the number of steps on the posttest, with higher optimality scores given for fewer steps, explained in detail below in section \ref{sec:pro_perf}.

We used only two of the datasets, f18 and s19 ($\mathit{N} = 437$), to perform our correlation analysis, since the problem ordering changed in Fall 2018. Using these two datasets allows us to control for differences arising from problem ordering. For each student, we computed their posttest optimality and the proportion of training steps that are labeled HelpNeed using each efficiency metric. We then calculated the correlation between each type of training HelpNeed with posttest optimality using Pearson’s coefficient. Table \ref{tab:corr} shows that training HelpNeed is significantly and negatively correlated to posttest optimality for all four of the efficiency metrics. A negative correlation suggests that the higher the proportion of HelpNeed steps in training, the worse the posttest optimality. Among the step efficiency metrics, the global-absolute metric is the most correlated with posttest optimality. Therefore, we selected the  \textbf{Global-Absolute step efficiency} to detect when students require tutor interventions. The remainder of the paper uses the term HelpNeed to refer to the definition that uses the global-absolute step efficiency.

\subsection{HelpNeed Predictor}
\label{sec:predictor} 

Our prior work on proactive hints suggests that low prior knowledge students exert productive persistence when the tutor provides proactive hints as partially worked steps  \cite{maniktala2020leveraging}. Therefore, we sought to build a predictive HelpNeed classifier that could identify unproductive steps at their start, so that we could proactively intervene with a hint, and possibly convert unproductivity to productive persistence. This HelpNeed predictor predicts two classes: 1 for predicting HelpNeed, and 0 for otherwise. 

To build the HelpNeed predictor, we engineered two types of classifiers: \textit{state-based}, and \textit{state-free} in Python. We use a \textit{state-based} classifier when a student's problem-solving state can be matched to historical data to leverage quality, and progress-based features for predictions (see Appendix \ref{aa}). The \textit{state-free} classifier is used when we don’t have that information. This two-classifier architecture ensures that a HelpNeed prediction can be made regardless of whether a state is present in the historical data.

\textbf{Feature Engineering and Selection}. We used the datasets f18, and s19 ($\mathit{N} = 437$)  to develop the HelpNeed predictor. We aggregate features on three levels of granularity: 1) most recent step, 2) current problem, and 3) total (all student-tutor interactions). Appendix \ref{aa} provides more information about on 63 features. We normalized the features and performed feature selection using the scikit-learn feature\textunderscore selection.SelectFromModel \cite{pedregosa2011scikit}. Appendix \ref{ab} shows the features selected, along with their descriptive statistics. 

Related research on creating data-driven step-level hint policies in multi-step domains for probability \cite{zhou2019hierarchical} and linked lists \cite{fossati2015data} have used individual models for each problem. However, we developed a generalized cross-problem model, and compared it with problem-specific models. The generalized model performed equally well in AUC and recall when compared to problem-specific models for most of our tutor problems. Further, the generalized model performed better than the problem-specific models when less historical data was available, confirming that using a generalized model is a reasonable approach. 

We now detail the predictive models we experimented with and our model selection approach.
Table \ref{tab:corr} shows that the input data is imbalanced as the proportion of HelpNeed steps is small at 24.86\%. So, we used 10-fold cross-validation with stratified random sampling\footnote{We also ensured that each student’s data was fully contained within either the training or test set of each fold to avoid introducing bias}. We experimented with nine HelpNeed classifiers including Random Forest (RF), Decision Tree (DT), Support Vector Classifier (SVC), Multi-layer Perceptron (MLP), Quadratic Discriminant Analysis, K-Nearest Neighbours, AdaBoost, Naive Bayes, all via scikit-learn, and XGBoost \cite{chen2016xgboost}. We used the \textit{sklearn} class\textunderscore weight to determine \textit{automated} weights for the RF, DT, SVC, and MLP models to account for imbalanced data. 
So the model would be more likely to correctly identify HelpNeed steps, we created \textit{expert} weights (to improve recall) by performing a grid search starting from the automated weights for class 0, and performed a grid search for class 1 (HelpNeed), scoring the models on both recall and AUC \cite{gridsearch}.

Table \ref{tab:recall_auc} shows the result of applying 10-fold cross validation on the classifiers with default, automated, and expert class weights for state-based and state-free predictions. We omitted the results for classifiers where class weights could not be adjusted because of their low performance. The table shows the RF models with expert class weights have the highest recall for both state-based and state-free predictions (0.90 and 0.91, respectively)\footnote{The automated weights for these RF models are \{class 1: 2.09, class 0: 0.66\} while the expert weight for class 1 is 4.17 ($\sim$2x) for the \textit{state-based} predictions and 3.75 ($\sim$1.8x) for the \textit{state-free} predictions.}. 
RF outperforms all other models because it is both an ensemble model and benefits from expert weights.


\begin{table}[]
\centering
\vspace{2em}
\caption{Comparison of state-based and state-free predictors: Recall and AUC for 10-fold Cross Validation; Dflt= Default model; Auto = using automated weights; Exp = using expert weights}
\label{tab:recall_auc}
\begin{tabular}{|c|c|c|c|c|c|c|c|c|c|c|c|c|}
\hline
\multirow{3}{*}{\begin{tabular}[c]{@{}l@{}}Predictor\end{tabular}} & \multicolumn{6}{c|}{State-based}                       & \multicolumn{6}{c|}{State-free}                                                                                                                                             \\ \cline{2-13} 
                            & \multicolumn{3}{c|}{Recall} & \multicolumn{3}{c|}{AUC} & \multicolumn{3}{c|}{Recall}                                                          & \multicolumn{3}{c|}{AUC}                                                             \\ \cline{2-13} 
                            & Dflt   & Auto   & Exp   & Dflt  & Auto  & Exp  & \multicolumn{1}{l|}{Dflt} & \multicolumn{1}{l|}{Auto} & \multicolumn{1}{l|}{Exp} & \multicolumn{1}{l|}{Dflt} & \multicolumn{1}{l|}{Auto} & \multicolumn{1}{l|}{Exp} \\ \hline
RF               & .62    & .83   & \color{blue} \textbf{.90}  & .71   & .84  & \color{blue} \textbf{.83} & .20                       & .51                      & \color{blue} \textbf{.91 }                   & .52                       & .62                      & \color{blue} \textbf{.62}                     \\ \hline
DT               & .72    & .84   & .89  & .82   & .84   & .82 & .22                       & .46                      & .79                     & .53                       & .60                      & .54                     \\ \hline
SVC                         & .58    & .52   & .63  & .77   & .68  & .73 & .12                       & .28                      & .42                     & .53                       & .52                      & .52                     \\ \hline
MLP                         & .64    & .62   & .44  & .75   & .74  & .67 & .08                       & .24                      & .32                     & .53                       &.55                      & .54                     \\ \hline
\end{tabular}
\end{table}

For the selected RF state-based classifier, quality- and progress-based features contribute to 94.81\% of the predictive power. Amongst these, the top three features to predict HelpNeed are: (1) Global-Absolute Progress (GAP = 33.5\%), (2) current state’s Global Quality = 22\%), and (3) Local-Absolute Progress (LAP = 13.1\%). And, for the selected RF state-free classifier, the top three features contributing to the classifier's predictive power are: (1) problem time (pTime = 10.7\%), (2) total click-based actions in a problem (pActionCount = 8.5 \%), and (3) wrong rule applications in the problem (pWrongApp = 7.4\%). These three features are likely to be available in most well-structured multi-step domains. In ill-structured domains, there may not exist a program that can detect whether a single rule application is correct or not. More information on feature importance is provided in Appendix \ref{ac}.

The HelpNeed predictor is defined as the combined state-based and state-free RF classifiers that predict the next step’s HelpNeed classification as in Table \ref{tab:def_unp} at the start of that step using the global-absolute metric. We performed a semester-based two-fold cross validation with f18 and s19 as the two folds to assess if the predictor can be used across semesters. We observed similar recall and AUC as that observed in the 10-fold cross validation, with recall for state-based predictions = 0.89 and state-free predictions = 0.89; and AUC for state-based predictions = 0.88 and state-free predictions = 0.56. This confirms that the HelpNeed predictive classifiers trained here can be expected to work in the future without the need for re-training.



\section{Experiment} 
We conducted a controlled experiment to compare our \textit{Adaptive} condition, where participants received proactive hints (Figure \ref{fig:assertion_justification}) when our predictor indicated HelpNeed and the \textit{Control} condition, where participants worked as usual without proactive hints. Note, students in both conditions could request on-demand hints. 

\subsection{Hypotheses}
We have two hypotheses:
\begin{itemize}
\item \textbf{H1}:  Students in the Adaptive condition will have better posttest performance than those in the Control condition, as measured by solution optimality and time.
\item \textbf{H2}: Students in the Adaptive condition will exhibit better training behaviors, with (a) fewer HelpNeed steps, and (b) lower possible help avoidance, and higher possible help appropriateness (a higher chance of receiving help when it was likely to be needed), as measured using the HelpNeed classifier, when compared to the Control. 
\end{itemize} 

\subsection{Procedure}
\label{sec:procedure}
The tutor was a homework assignment in a Fall 2019 undergraduate discrete math course. The study was conducted with 123 participants, and the students were given ten days to complete the tutor.

The tutor is divided into four sections: introduction, pretest, training, and posttest. The introduction presents two worked examples to familiarize students with the tutor interface. Next, students solve two problems in a \textit{pretest}, which is used to determine students’ incoming competence. Students are assigned a condition based on their pretest performance for stratified sampling, as detailed below. The pretest problems are designed to be easy and short, solvable with short optimal solution lengths ($\mathit{Mean} = 3.5$, $\mathit{SD} = 0.71$). Next, the tutor guides students through the \textit{training} section with fifteen problems of varying difficulty. The difficulty of the training problems is between that of the pretest and the posttest  based on averaging the optimal solution lengths over all training problems ($\mathit{Mean} = 4.99$, $\mathit{SD} = 1.32$). Finally, students take a more difficult \textit{posttest} with five problems, with longer optimal solution lengths compared to the other sections ($\mathit{Mean} = 7.25$, $\mathit{SD} = 1.89$). Note that students can only receive hints in the training but the tutor is designed to provide immediate feedback on rule application errors in all the sections. 

A stratified random sample based on pretest performance is used to partition the 123 participants into two conditions, resulting in 70 in \textit{Adaptive} and 53 in \textit{Control}. The stratified sampling was set up to result in a larger sample size for the Adaptive condition to gather more data on how the intelligent policy was carried out. Among these participants, 111 (66 in Adaptive and 45 in Control) completed the tutor. We used a chi-squared test to assess the impact of tutor completion rates on the group sizes, and found that the impact was not significant ($\mathit{\chi^2}(1, \mathit{N} = 123) = 0.16,  \mathit{p} = .69$). 

\subsection{Performance Metrics}
\label{sec:pro_perf}
The tutor automatically checks for problem completeness, so student solutions cannot be incorrect, but some may be more expert than others. Learning is measured by \textit{performance} measures of optimality, time, and accuracy, reflecting that expert-like problem solutions have fewer steps, take less time, and have fewer mistakes. \textit{Optimality} is an exponential decay function on normalized steps $e^{-steps}$ to account for the small variance in the number of steps.
$Steps$ are normalized to the interquartile range for each specific problem to account for varying problem lengths. Very short solutions with steps less than or equal to Q1 ($1^{st}$ quartile) have $optimality = 1$, and those with steps greater than Q3 ($3^{rd}$ quartile) have an optimality score of $0.36$ or less based on the exponential decay curve. Next, similar to other studies \cite{kardan2015providing,tchetagni2002hierarchical}, our performance includes the \textit{time} students spend solving problems. To make the time data into approximate normal distributions, we cap each action (any click performed) time to one minute\footnote[2]{The $3^{rd}$ quartile of action time in Fall 2019 was 4.6s, and only 4690 out of 277,647 actions had an action time greater than 1 minute}, and sum the times for each action to determine the total (capped) time per problem. Finally, \textit{accuracy} is defined as the number of correct rule applications divided by the total rule applications. 

We do not hypothesize differences in the accuracy between the two conditions because the tutor is designed to provide immediate feedback on incorrect rule applications without penalties, even within the pre- and post-tests (see \ref{sec:procedure}). We report this performance metric to ensure our intervention does not harm students’ accuracy.

\subsection{Hint Usage}
\begin{figure*}
\centering
\includegraphics[width=0.8\columnwidth]{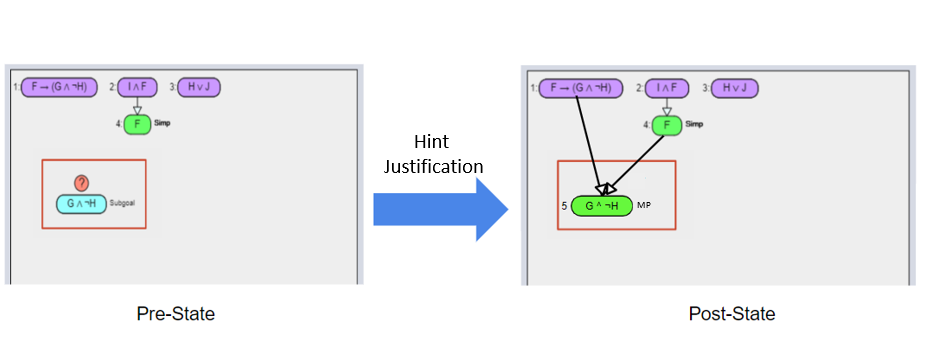}
\caption{Proactive hint justification: A cyan proactive hint node (labeled Subgoal) is shown in the pre-State on the left. ``Hint Justification” is performed by clicking on nodes 1 and 4 and the MP rule button in the pre-state. Since this is a correct justification, the hint becomes incorporated as a green node in the solution as shown in the post-state on the right.}
\label{fig:assertion_justification}
\end{figure*}

Since our study investigates proactive hints, it is imperative to understand students’ hint usage. We measure hint usage in the tutor using the Hint Justification Rate (HJR). As shown in Figure \ref{fig:assertion_justification}, a hint becomes \textit{justified} if a student selects the correct rule and existing nodes to derive it. HJR is defined as the proportion of the hints given (on-demand or proactive) that were justified.

\section{Results}
In this section, we investigate our two hypotheses about the Adaptive condition: H1 on improved posttest performance, and H2 on (a) reduced HelpNeed steps, and (b) reduced possible help avoidance and increased possible help appropriateness during training. We also evaluate the predictor’s efficacy in predicting HelpNeed, and show how the predictor can be used to assess students’ help-seeking behavior.

\subsection{H1 - Posttest Performance} \label{posttest}
We first investigate hypothesis H1 on whether the Adaptive students have better posttest optimality and time than their Control peers. Table \ref{tab:posttest} shows the distribution parameters of students’ pre-, and post-test performance in the two Conditions \{Adaptive, Control\}. As expected, no significant differences were found between the two conditions in the pretest \textit{optimality}: \textit{t}(110) = 0.93, \textit{p} = .18, \textit{time}: \textit{t}(110) = 1.27, \textit{p} = .11, and \textit{accuracy} (proportion of correct rule applications): \textit{t}(110) = 0.81, \textit{p} = .37. This confirms that our stratified random sampling assignment balanced Adaptive vs. Control conditions' incoming competence (see section \ref{sec:procedure}).

\begin{table}[]
\centering
\caption{The distribution parameters for students’ pre- and post-test performance in the two conditions. T-test shows that the Adaptive group has significantly higher posttest optimality (\textit{p} = .04), and took significantly less time to complete the posttest (\textit{p} $<$ .01) than the Control group}
\begin{tabular}{|c|c|c|c|c|c|c|}
\hline
\multirow{2}{*}{Test} & \multicolumn{2}{c|}{Optimality} & \multicolumn{2}{c|}{Time (min)} & \multicolumn{2}{c|}{Accuracy} \\ \cline{2-7} 
                      & Adaptive       & Control        & Adaptive    & Control     & Adaptive      & Control       \\ \cline{1-7} 
Pretest               & .54 (.38)    & .60 (.27)    & 39 (20) &  34 (16) & .62 (.12)   & .62 (.14)   \\ \hline
Posttest              & \color{blue}.71 (.27)    &\color{blue} .59 (.33)    & \color{blue} 18 (12) &\color{blue} 29 (17) & .71 (.10)   & .68 (.08)   \\ \hline
\end{tabular}
\label{tab:posttest}
\end{table}

A t-test on the posttest optimality shows a significant difference between the two conditions (\textit{t}(110) = 1.74, \textit{p} = .04) with a moderate effect size (Cohen's d = 0.4). Recall that higher optimality values reflect shorter, more efficient solutions. Next, on the total posttest time\footnote{The posttest time distributions were normal but there was a significant difference in variance between the two groups using the Levene’s Test (\textit{p} = .02), so, we conducted a Welch’s t-test to test for significant differences on post-test time between conditions.}, significant differences were found between the two conditions with a large effect size (\textit{t}(110) = 3.99, \textit{p} $<$ .01, Cohen’s d = 0.8s), with students in the Adaptive condition (\textit{M} = 18 min, \textit{SD} = 12 min) spending significantly less time on the posttest than those in the Control (\textit{M} = 29 min, \textit{SD} = 17 min). No significant differences were hypothesized or found  between the two conditions in the posttest accuracy (\textit{t}(110) = 0.50, \textit{p} = .31). These results show that delivering proactive hints using the HelpNeed predictor indeed helped the Adaptive students generate \emph{better} solutions in \emph{less} time than their Control peers on the posttest without negatively impacting their accuracy, confirming our first hypothesis H1. 

\subsection{H2a - Comparison of HelpNeed during training} 
\label{train}

\begin{table}[]
\centering
\caption{Distribution parameters for students’ training steps in the two conditions. T-tests show significant differences between the two conditions in the Far Off (\textit{p} $<$ .01) and Opportunistic (\textit{p} = .02) steps. Total training steps are only marginally significant (\textit{p} = .10)}
\begin{tabular}{|l|l|c|c|c|}
\hline
\multicolumn{1}{|c|}{}                                & \multicolumn{1}{c|}{}                                     & \multicolumn{2}{c|}{\# Training Steps}                              &                               \\ \cline{3-4}
\multicolumn{1}{|c|}{\multirow{-2}{*}{Step Behavior}} & \multicolumn{1}{c|}{\multirow{-2}{*}{Description}}   &  Adaptive                   & Control                    &   \multirow{-2}{*}{\textit{p}}                            
\\ \hline
Expert                                          & Quick efficient steps                                     & 61 (12)                        & 65 (13)                        & .10                         \\ \hline
Strategic                                             & Long efficient steps                                       & 25 (19)                        & 21 (9)                         & .17                         \\ \hline
{\color[HTML]{3531FF} Opportunistic}                     & {\color[HTML]{3531FF} Singular, quick, inefficient steps}           & {\color[HTML]{3531FF} 5 (3)}   & {\color[HTML]{3531FF} 7 (4)}   & {\color[HTML]{3531FF} $<$.01*} \\ \hline
{\color[HTML]{3531FF} Far Off}                  & {\color[HTML]{3531FF}  Consecutive quick but inefficient steps} & {\color[HTML]{3531FF} 16 (20)} & {\color[HTML]{3531FF} 25 (25)} & {\color[HTML]{3531FF} .02*} \\ \hline
Futile                                                & Long inefficient steps                                   & 13 (12)                        & 12 (19)                        & .47                         \\ \bottomrule
\multicolumn{2}{|r|}{Total Training Steps}                                                                        & 121 (38)                       & 133 (39)                       & .10                         \\ \bottomrule
\end{tabular}
\label{tab:training_steps}
\end{table}
In this section, we investigate hypothesis H2a that the students in the Adaptive condition will have significantly fewer HelpNeed steps in training than the Control condition. Table \ref{tab:training_steps} shows the cumulative step-level behavior of the two conditions during training. A t-test on the total training steps shows that the Adaptive condition took marginally significantly fewer total training steps on average compared to the Control condition: \textit{t}(110) = 1.29, \textit{p} = .10. The Adaptive condition also has significantly fewer quick inefficient, Far Off and Opportunistic, steps per student over all training problems than the Control  (Opportunistic - Adaptive: 5, Control: 7, \textit{p} $<$ .01, and Far Off - Adaptive: 16, Control: 25, \textit{p} = .02) but there were no significant differences in Futile steps between the two conditions (Adaptive: 13, Control: 12, \textit{p} = .45).  This suggests  that compared with the Control condition, the Adaptive condition avoided unnecessary Opportunistic and Far Off steps that might distract them away from efficient solutions. Table \ref{tab:training_steps} also shows that there are no significant differences in the Expert or Strategic steps between the two conditions (Expert: \textit{p} = .10, Strategic: \textit{p} = 0.17). The significantly higher Opportunistic and Far Off steps in the Control condition may be a result of help avoidance because students may not know when to seek help \cite{pena2011improving,azevedo2004does}. More details on students’ help avoidance are discussed in subsection \ref{sec:eval_help}.

\subsection{Hints Given and Used in Training}
\label{hint_use}

In this section, we further investigate the sources of differences between the Adaptive and Control conditions. Table \ref{tab:hint_usage} shows the mean and standard deviation of the total number of proactive, on-demand, and overall hints received by all students in the two conditions across training problems (the top part), and the hint justification rate (HJR) of students in the two conditions across different types of hints (the bottom part). Note that the Control condition was not provided with proactive hints, and thus only their on-demand hints count toward their total hints. 

For the Adaptive condition, the total hints include both proactive and on-demand hints (Mean = 34, SD = 12), whereas, for the Control condition, the total hints only include the on-demand hints (Mean = 14, SD = 15). A Mann Whitney U test on the total hints for the two conditions shows a significant difference (\textit{U} = 445.5, \textit{z} = 6.24, \textit{p} $<$ .01). On average, students in the Adaptive condition received 28 proactive hints (22.8\% of steps), and 6 on-demand hints (4.9\% of steps) while the Control condition received an average of 14 hints on-demand (10.5\% of steps). A Mann Whitney U test shows that students in the Adaptive condition requested significantly fewer on-demand hints during training than those in the Control condition (\textit{U} = 908.5, \textit{z} = 3.46, \textit{p} $< .01$). This suggests that our HelpNeed predictor and proactive hints may have successfully forecasted the Adaptive students’ needs for hints. Interestingly, we also found that whenever students in the Adaptive condition requested hints in a step, they did so after working on the step a median of 44 seconds before seeking help (Mean = 70s, SD = 125s), whereas the Control condition requested hints after working on the step for a median of  21 seconds (Mean = 29s, SD = 21s). A Mann Whitney U test on the time spent in a step before requesting help shows a significant difference between the two conditions (\textit{U} =  600, \textit{z} = 1.97, \textit{p} = .02). This suggests that on average, students in the Adaptive condition may engage longer on their own than their Control peers before requesting hints on a step. More details on help abuse are discussed in subsections \ref{sec:game}, and \ref{sec:eval_help}.

\begin{table}[]
\centering
\caption{Distribution parameters for the number of hints given and the hint justification rate in the two conditions}
\begin{tabular}{cc|c|c|c|}
\cline{3-5}
\multicolumn{1}{l}{}                                                                                          & \multicolumn{1}{l|}{} & Adaptive      & Control                  &                              \textit{p} \\ \hline
\multicolumn{1}{|c|}{}                                                                                        & Proactive            & 28 (10)  & - &-     \\ \cline{2-5} 
\multicolumn{1}{|c|}{}                                                                                        & On-demand             & 6 (9)   & 14 (15)            & \textless{}.01*              \\ \cline{2-5} 
\multicolumn{1}{|c|}{\multirow{-3}{*}{\begin{tabular}[c]{@{}c@{}} \# Hints \\ Received\end{tabular}}}             & Total Hints                & 34 (12) & 14 (15)            & \textless{}.01*              \\ \hline
\multicolumn{1}{|c|}{}                                                                                        & Proactive            & 89\% (8\%)  & - & -     \\ \cline{2-5} 
\multicolumn{1}{|c|}{}                                                                                        & On-demand             & 87\% (22\%) & 90\% (11\%)            & .23                        \\ \cline{2-5} 
\multicolumn{1}{|c|}{\multirow{-3}{*}{\begin{tabular}[c]{@{}c@{}}\% Hint \\ Justification \\ Rate (HJR) \end{tabular}}} & Total HJR                & 89\% (8\%)  & 90\% (11\%)            & .26                        \\ \hline
\end{tabular}
\label{tab:hint_usage}
\end{table}


For HJR, the total hints HJRs are high, around 90\% for both conditions, and no significant difference was found between the two conditions (\textit{U} = 1074.5, \textit{z} = 1.21, \textit{p} = .11), which suggests that our hints are well-accepted by students in both conditions. More specifically,  no significant differences are found in the on-demand HJR between the two conditions (Means: Adaptive: 87, and Control: 90, \textit{U} = 513.5, \textit{z} = 1.84, \textit{p} =  .03). Further, the Adaptive condition justified most of their hints regardless of delivery type (proactive HJR: 89\%, and on-demand HJR: 87\%). This affirms our prior results that students in the Adaptive condition incorporated proactive hints into their solutions as frequently as on-demand hints  \cite{maniktala2020leveraging}.

\subsection{Hint Count and Posttest Performance}
\label{sec:hintcorr}
We found that the Adaptive condition received significantly more hints during training (subsection \ref{hint_use}) and also performed significantly better on the posttest in optimality and time (subsection \ref{posttest}). We perform a correlation analysis comparing the Adaptive condition to students who also received frequent proactive hints at random times in previous semesters to determine whether this difference in performance could be due to the increased number of hints provided to the Adaptive group. Ideally, we would compare the HelpNeed policy with a policy that provides the same number of proactive hints randomly. However, this is difficult to achieve, since the policy is adapting to individual students and the total number of proactive hints per student is neither predetermined nor consistent. Since we cannot directly compare student performance because of varying hint frequencies, we conducted a correlation analysis instead. Further, we contrast the correlation of hint count with posttest performance while controlling for pretest performance across datasets with Random, a combined dataset with Random and Adaptive proactive hint conditions, and the Adaptive condition alone. This can help us to understand differences in correlation arising from the Adaptive condition. 

Table \ref{tab:corr_hint_perf} shows the correlation between hint count (total and proactive) and posttest performance (Optimality, and Time) with pretest performance as the covariates. First, we analyzed prior datasets, where students were given proactive hints at random intervals with about 40\% and 33\% frequency in f18R and f19R\footnote[1]{f18R,  and s19R were collected in Fall 2018 and Spring 2019 semesters respectively where R represents a subset of the population that received frequent unsolicited hints provided randomly} respectively. Proactive hints in f18R, and s19R were given frequently at random times irrespective of prior knowledge or progress on the problem, so \textit{within} each random condition, there was no significant difference in the proactive hint count between students with low and high prior knowledge \cite{maniktala2020leveraging}. 
For the f18R + s19R dataset, we found that the total hints (proactive + on-demand) did not correlate to the posttest performance metrics. Further, the number of proactive hints significantly correlated to longer posttest time but did not significantly correlate to posttest optimality. This suggests that receiving higher numbers of proactive hints on random steps during training can be detrimental to students’ posttest time.  

Proactive hints were provided on average of 23\% student steps in the f19 Adaptive condition (f19A) of this study, which is lower than the f18R and s19R proportions. We repeated the correlation analysis including the f19 Adaptive condition (f19A) of this study to check if the lower f19A proactive hint count led to better posttest performance than the higher f18R and s19R proactive hint counts. On this combined dataset, the total hints received was not significantly correlated with posttest performance. However, the higher number of proactive hints (that presumably occurred more often in the studies with more frequent random hints) significantly correlated with lower optimality and longer time. Finally, we performed the correlation between hint count and posttest performance for f19A alone. On this dataset, we observed insignificant correlations for all pairs of hint type \{total, proactive\}, and posttest metrics \{optimality, time\}. This is as expected because proactive hints were provided adaptively based on individual student needs, and led to improved performance for all.

\begin{table}[]
\caption{Correlation analysis between hints given and posttest performance metrics}
\label{tab:corr_hint_perf}
\centering
\begin{tabular}{|l|r|l|l|l|l|}
\hline
\multirow{2}{*}{Dataset}                                                     & \multicolumn{1}{c|}{\multirow{2}{*}{\begin{tabular}[c]{@{}c@{}}Hints\\ Given\end{tabular}}} & \multicolumn{2}{c|}{Optimality}                    & \multicolumn{2}{c|}{Time}                          \\ \cline{3-6} 
 & \multicolumn{1}{c|}{} & \multicolumn{1}{c|}{corr} & \multicolumn{1}{c|}{p} & \multicolumn{1}{c|}{corr} & \multicolumn{1}{c|}{p} \\ \hline
\multicolumn{1}{|l|}{\multirow{2}{*}{f18R, s19R}}                            & Total & -0.05  & .62 & 0.16 & .13  \\ \cline{2-6} 
\multicolumn{1}{|l|}{}  & Proactive & -0.14 & .19 & \color[HTML]{3531FF}0.49 & $<$ \color[HTML]{3531FF}.01 \\ \hline
\multirow{2}{*}{\begin{tabular}[c]{@{}c@{}}f18R, s19R, f19A\end{tabular}} & Total  & -0.05 & .52 & 0.13 & .11 \\ \cline{2-6} 
& Proactive & \color[HTML]{3531FF}-0.17 & \color[HTML]{3531FF}.04 & \color[HTML]{3531FF}0.27  & \color[HTML]{3531FF} $<$ .01 \\ \hline
\multirow{2}{*}{\begin{tabular}[]{@{}c@{}}f19A\end{tabular}} & Total  & -0.003 & .98 & -0.15 & .24 \\ \cline{2-6} 
& Proactive & 0.15 & .24 & 0.04  & .77 \\ \hline
\end{tabular}
\end{table}

This analysis suggests that providing too many proactive hints can be detrimental to student performance. These results imply that students \textit{do} need more hints than the number they request, but providing hints when they are needed is more important than simply providing more.

\subsection{Gaming behavior}
\label{sec:game}
Since studies suggest that some students may game an educational system to complete the problems faster by using hints frequently \cite{baker2009educational}, we analyze gaming behaviors. Our Adaptive hints policy is complex, so students are unlikely to guess what causes proactive hints to be triggered. A student trying to game the proactive hints in our tutor would have to carry out Futile or Far Off steps to receive proactive hints. However, Table \ref{tab:training_steps} (training steps) shows that the Adaptive condition students had significantly fewer Far Off and Opportunistic steps during training than Control, and a similar number of Futile steps, giving no indication of increased gaming. Further, 
proactive hints that are partially worked steps require the student to do some work to use them, reducing the potential for gaming the system, or possibly turning such attempts into learning opportunities.

\begin{figure*}
\centering
\caption{Comparison of Gaming in Deep Thought between the two conditions}
\begin{subfigure}{.49\textwidth}
  \centering
  \includegraphics[width=\linewidth]{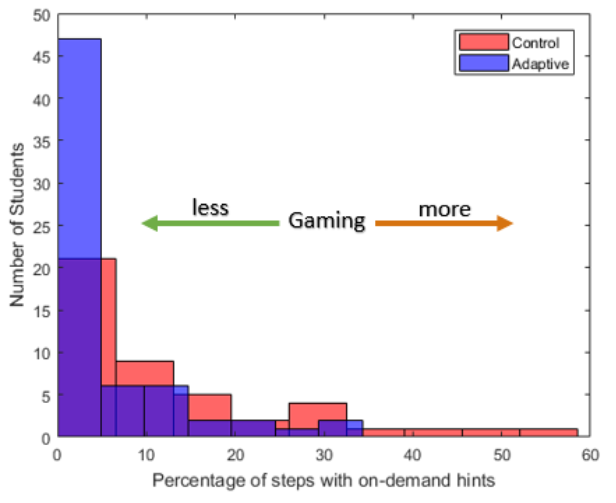}
  \caption{Histogram for the proportion of training \\steps with hint requests}
 \label{fig:on_dem_perc}
\end{subfigure}
\begin{subfigure}{.49\textwidth}
  \centering
  \includegraphics[width=\linewidth]{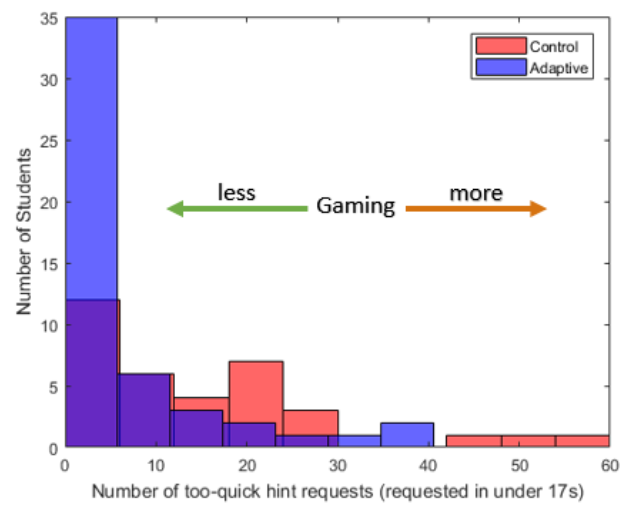}
  \caption{Histogram for the number of too-quick hint \\requests}
  \label{fig:quick_req}
\end{subfigure}
\label{fig:gaming}
\end{figure*}

We also analyzed how often students were gaming the system by requesting more hints than needed. Figure \ref{fig:on_dem_perc} shows that the on-demand hints in both the conditions constituted a small proportion of each student's total steps (Adaptive: Mean = 4\%, SD = 8\%; Control: Mean = 12\%, SD = 14\%). We found only one student from the Control condition who requested hints on more than 50\% of their steps. Next, we investigated how often students requested hints in a short step time, where we consider “short” to be the top 25\% quickest requests. We found that the $25^{th}$ percentile ($1^{st}$ quartile) of step time before a student requested a hint in the study is 17 seconds. Figure  \ref{fig:quick_req} shows the histogram of students in the two conditions, comparing the number of training steps with too-quick hint requests (requested under 17 seconds). Summed over all problems in the training section, each student in the Adaptive condition had significantly fewer too-quick hint requests (requested in under 17s) than those in the Control condition (Adaptive: Mean = 6 too-quick requests per student, SD = 10; Control: Mean = 15 too-quick requests per student, SD = 15; \textit{U} = 498.5, \textit{z} = 3.36, \textit{p} $<$ .01). This suggests that students in the Adaptive condition are less likely to game the system by requesting hints without engaging long enough to think about the next step, than students in the Control condition.

Overall, we observed that students in the Adaptive condition were less likely to game proactive or on-demand hints, with significantly fewer quick inefficient steps. The design of the proactive hints seems to successfully prevent gaming behaviors that detract from learning. However, we provide more details on help abuse in the next two sections.

\subsection{The Policy’s Efficacy in Predicting HelpNeed }
\label{sec:viz}
In this section, we evaluate the HelpNeed predictor by comparing the HelpNeed predictions with the observed step behavior during training for both conditions. Note that this comparison is complex, because students can request hints at any time, and a hint received (proactive or on-demand) can change a step’s observed HelpNeed if and when it is justified. Comparing the performance of our HelpNeed predictor on the two conditions can provide us some insights into how the predictor performs with and without predictor-driven interventions. We also use the predictor to compare student help-seeking behavior between the two conditions.

We evaluate the performance of the predictor in a manner similar to a confusion matrix, comparing step predictions with observed step behavior on 7970 and 5975 steps respectively for the Adaptive (\textit{N} = 66 students), and Control (\textit{N} = 45 students) conditions\footnote{The number of steps is larger for the Adaptive condition since there are more students assigned to this condition.}. As an exploratory analysis, we compared the total HelpNeed predictions and observed HelpNeed steps between the two conditions. If a step is classified as HelpNeed, we call it an \textit{observed HN} step. Alternatively, a step that is not classified as HelpNeed is called an \textit{observed OK} step. Similarly, if the HelpNeed predictor predicts HelpNeed in a step, we call it a \textit{predicted HN} step, otherwise, we call it a \textit{predicted OK} step. We found 1858 (23.3\%) observed HN training steps for the Adaptive group, and 1660 (27.8\%) observed HN training steps for the Control group. A chi-square test on the observed HN vs. OK training steps between the two conditions shows a significant difference ($\mathit{\chi^2}(1, \mathit{N} = 111) = 36.2,  \mathit{p} < .01$). The Adaptive condition has a percent HelpNeed steps only 4.5\% lower than that for the Control because the Adaptive condition’s total training steps are also lower (marginally significant) than that of the Control. Section \ref{train} shows that the Adaptive condition has significantly fewer Far Off and Opportunistic steps than the Control, which contributes to their reduced total steps as well. So, even though the Adaptive condition has fewer HelpNeed steps, their lower total steps make the difference in percent-HelpNeed between the two conditions small. We also compared the predictions of HelpNeed between the two groups. Interestingly, the Control condition had a noticeably higher proportion of predicted HN training steps than the Adaptive condition (Control: 2296, 38.4\%, Adaptive: 1820, 22.8\%). A chi-square test on predicted HN vs. OK training instances between the two conditions shows a significant difference ($\mathit{\chi^2}(1, \mathit{N} = 111) = 399.0,  \mathit{p} < .01$). This result is consistent with the findings that students in the Adaptive condition were likely to follow their frequent hints and this resulted in a higher proportion of observed OK steps, leading to fewer HN predictions.

Next, we investigated the correctness of step predictions and the impact of hint provision on the observed step behavior. Figure \ref{fig:stack_viz} shows bar graphs comparing training steps in the two conditions on whether a step is observed OK (top) or observed HN (bottom). Furthermore, each step can be categorized as  prediction of HN/OK steps and hinted/no-hints (including both on-demand or proactive)\footnote{Note, the graph provides only the percentages. The numbers corresponding to these eight comparisons are presented in Appendix \ref{conf_mat}}. Therefore, for both observed OK or observed HN steps, the proportions of training steps are further categorized based on the combination of HelpNeed predictor results \{pred-OK = predicted OK, or pred-HN = predicted HelpNeed\} with  Hint provision \{noHint = no hint was provided, hinted = a hint was provided\} which resulted in four types: 

 \begin{figure}[h!]
\centering
\includegraphics[width=0.80\columnwidth]{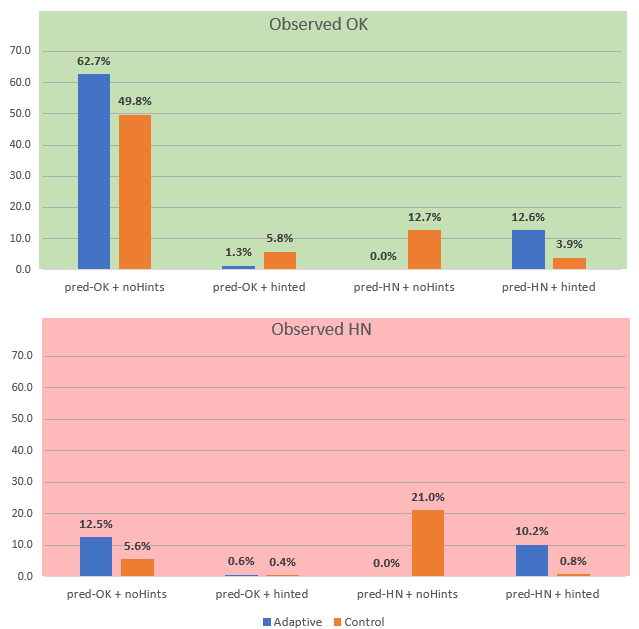}
\caption{Comparing training steps in the two conditions on a combination of three aspects: predictions of HN or OK, observed HN or OK, and hint provision (on-demand or proactive)}
\label{fig:stack_viz}
\end{figure}

\begin{enumerate}[noitemsep]

\item  \emph{pred-OK + noHints}: predicted OK and no hints were provided
\item  \emph{pred-OK + hinted}: predicted OK but hints were provided
\item  \emph{pred-HN + noHints}: predicted HelpNeed but no hints were provided
\item  \emph{pred-HN + hinted}: predicted HelpNeed and hints were provided 
\end{enumerate}

The observed OK steps are shown in the top portion of Figure 6. The Adaptive group (4998, 62.7\%) has a much higher percentage of \textbf{\emph{pred-OK + noHints}} steps than the Control group (2974, 49.8\%). This suggests that, while both conditions can solve a large number of steps on their own, the Adaptive students can successfully solve more steps on their own than the Control. On \textbf{\textit{pred-OK + hinted}} steps, interestingly, we observe a very small proportion of such steps for the Adaptive group (106, 1.3\%), considerably lower than that for the Control group (348, 5.8\%). Assuming our HelpNeed predictor was perfect, these numbers would reflect the possible help abuse steps where students asked for hints when they didn't need them. While it is possible that our HelpNeed predictor could have incorrectly predicted these steps to be OK, we believe that these numbers still provide insights into relative possible help abuse between different conditions. For \textbf{\textit{pred-HN + noHints}} steps, no such steps were observed for the Adaptive group simply because proactive hints were automatically provided for all predicted HN steps. However, we observe 761 (12.7\%) such instances for the Control condition, where the HelpNeed predictor incorrectly predicted a step to be HN but was observed to be OK without help. Ideally we want the misclassification rate to be as low as possible, with both the rates of false positives (predicted HN and observed OK) and false negatives (predicted OK and observed HN) low. However, we trained the predictors to improve recall (as discussed in section \ref{sec:predictor}) by lowering the rate of false negatives, i.e., we trained the model to prioritize correctly predicting HN steps so we could provide hints when they were most needed. Finally, for \textbf{\textit{pred-HN + hinted}} steps, as expected, we observe a higher proportion of \textit{pred-HN + hinted} steps for the Adaptive group (1008, 12.6\%) than the Control group (232, 3.9\%). We expected this number would be lower for students in the Control condition, who we thought would avoid asking for help. There are two possible scenarios for such Observed OK steps in the Adaptive group where students were predicted to need help, and received a hint: ideally, 1) the hint helped students achieve a more efficient step than predicted, or 2) a student was given help unnecessarily.

 Next, we examine the observed HN steps where steps were classified in a posthoc manner as Far Off or Futile. The lower portion of Figure \ref{fig:stack_viz} shows that for 
\textbf{\emph{pred-OK + noHints}} steps, where the predictor failed to identify HelpNeed, is higher for the Adaptive group (995, 12.5\%) than the Control group (335, 5.6\%). This result is likely because the predictor is strongly biased toward efficient steps present in student solutions, and proactive hints always suggest efficient steps. Therefore, when students used them, their attempts appeared more expert, which, in turn, caused the predictor to predict that their next steps would also be more expert. This suggests that our predictor needs revision to include hint usage since some expert-like and strategic behaviors by the Adaptive students may be due to previously-received, justified hints. The proportion of training steps in the next \textbf{\textit{pred-OK + hinted}} category is low in both conditions (Adopted: 51, 0.6\% and Control: 22, 0.4\%). The next category includes 1257 (21.0\%) \textbf{\textit{pred-HN + noHints}} steps for students in the Control condition, that were predicted and observed to need help, but no hints were received, representing possible help avoidance steps that the predictor correctly identified. 
Finally, the \textbf{\textit{pred-HN + hinted}} steps were predicted to need help, and received a hint, but still resulted in HelpNeed. The Adaptive group had 812 (10.2\%) and the Control group had 46 (0.8\%) of these steps. Note that our hints provide partial information to carry out the next step but a student may not use the hint immediately-- they may wait. Therefore, we further analyzed the usage of proactive hints in the Adaptive condition for this category and found that in 613 of these instances (75.5\% of  \textit{pred-HN + hinted-HN}), students justified the proactive hints in the second step after the hint was provided. This suggests that the proactive hints did help students in these steps, but it took two steps before we could observe this impact.

\subsection{Evaluating Help Behavior with HelpNeed predictor} \label{sec:eval_help}
In this section we investigate H2b, that students in the Adaptive condition would have lower possible help avoidance, and higher possible appropriate help, as measured using the HelpNeed classification, when compared to the Control. We also compare possible help abuse to determine whether the Adaptive hints impacted gaming behaviors. Note, we add a prefix \textit{possible} to these behaviors because HelpNeed does not represent ground truth as classified by experts. Rather, HelpNeed is our classification of steps needing help, and the predictor is a heuristic measure. 

\begin{table}
\centering
\caption{Definition of possible help avoidance, abuse, and appropriateness using HelpNeed}
\label{tab:haaa-def}
\begin{tabular}{|l|l|} 
\hline
Help Behavior Metric                          & Definition                                                                                 \\ 
\hline
Possible
  Help Avoidance       & \begin{tabular}[c]{@{}l@{}}\%steps with observed HelpNeed \\but no hints were 
  requested or received\end{tabular}    \\ 
\hline
Possible
  Help Abuse           & \begin{tabular}[c]{@{}l@{}}\%steps with no
  predicted or observed HelpNeed \\but hints 
  were requested\end{tabular}    \\ 
\hline
Possible
  Help Appropriateness & \begin{tabular}[c]{@{}l@{}}\%steps with
  predicted HelpNeed \\and hints were received\end{tabular}                    \\
\hline
\end{tabular}
\end{table}
 \begin{figure}
\centering
\includegraphics[width=0.70\columnwidth]{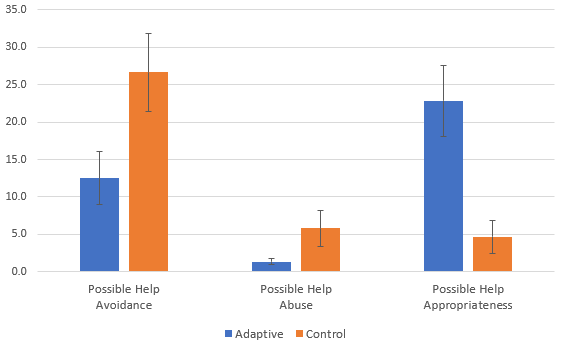}
\caption{Comparison of possible help avoidance, abuse, and appropriateness between conditions using our HelpNeed classification and predictor}
\label{fig:helpseeking}
\end{figure}

Table \ref{tab:haaa-def} provides the definitions for the help behaviors we investigated, and Figure \ref{fig:helpseeking} shows a comparison in these behaviors between the conditions, evaluated using the predictor. For each student, we define \textit{possible help avoidance} as the percentage of total training steps that were observed to be HelpNeed but hints (on-demand or proactive) were neither requested nor proactively provided. Figure \ref{fig:helpseeking} shows that the Adaptive condition has a mean of 12.5\% possible help-avoidance per student in training problems (SD = 3.5\%), and the Control condition has a mean of 26.6\% possible help-avoidance per student in training problems (SD = 5.2\%). We found significantly lower possible help avoidance in the Adaptive condition than the Control (\textit{U} = 138, \textit{p} $<$ .01) with a moderate effect size = 0.59\footnote{Effect size for Mann Whitney U test is calculated using $\dfrac{z^2}{n-1}$}. 
Next, we define \textit{possible help abuse} as the percentage of training steps with neither predicted nor observed HelpNeed, but students requested hints, indicating either that the prediction was wrong and help was needed and effectively used, or it was right but help was abused. We found a significant difference between the two conditions (\textit{U} = 365, \textit{p} $<$ .01) and a moderate effect size = 0.40, with the Adaptive condition (Mean = 1.3\%, SD = 0.4\%) having lower possible help abuse than the Control (Mean = 5.8\%, SD = 2.4\%).

Finally, \textit{possible help appropriateness} is the percentage of training steps predicted to need help and a hint was either requested or provided proactively.  Figure \ref{fig:helpseeking} shows that the Adaptive condition has a mean of 22.8\% possible help-appropriateness per student in training problems (SD = 4.8\%), and the Control condition has a mean of 4.7\% possible help-appropriateness per student in training problems (SD = 2.2\%). We found significantly higher possible help appropriateness in the Adaptive condition than the Control (\textit{U} = 155, \textit{p} $<$ .01) with a moderate effect size = 0.58. These results confirm hypothesis H2b that students in the Adaptive condition had lower possible help avoidance, and higher possible appropriate help than students in the Control. We further show that students in the Adaptive condition demonstrated lower possible help abuse than students in the Control.

\section{Discussion}
\label{sec:discussion}
In this paper, we provide a unique extension of the Hint Factory to determine productivity on a step-level in an intelligent data-driven tutor. We present a HelpNeed metric and predictor to identify steps where students are likely to need help learning efficient problem-solving strategies. We also show the analysis evaluating the impact of intervening with Adaptive hints using our HelpNeed predictor in a controlled study. In this section, we discuss the HelpNeed approach, the evaluation of the Adaptive hint policy, as well as its limitations and potential for generalization.

\subsection{The HelpNeed Metric and Predictor}
\label{sec:hn_disc}
Our definition of unproductive, HelpNeed behavior is different from the existing literature that either defines problem-completeness-based metrics \cite{beck2013wheel}, or uses pre-defined domain-specific metrics that require expert knowledge or domain modeling \cite{mclaren2014web}. We instead focus on solution length (i.e. optimality), which is valued across problem-solving domains. Further, our definition of HelpNeed also takes into account step duration. This is important because we do not want to disrupt helpful guess-and-check patterns that might involve students deriving an inefficient step but learning in the process. 

A study by Fossati, et al. on a proactive hint policy for the iList linked list tutor, incorporated step duration, using both problem-based parameters and individual student characteristics \cite{fossati2015data}. Similarly, we combine a problem-based parameter for a duration threshold after which steps are considered long, and each student's actual step duration to detect HelpNeed. The HelpNeed design ensures that we do not harm students who take, or need, more time when carrying out a learning task. Such students are likely to carry out longer steps that are classified as Strategic or Futile steps more than the quicker Expert, Far Off, or Opportunistic steps during training. Since we consider Strategic (long but efficient) steps productive, it gives such deliberating students more opportunities to learn by themselves. Further, a Futile (long and inefficient) step, indicates that the student could  benefit from proactive intervention, saving them time and directing them towards efficient strategies. Our prior work suggests that proactive hints promote learning through self-explanation and foster productive persistence among students with low prior knowledge (e.g. those with long step durations in the pre-test)  \cite{maniktala2020leveraging}. While this HelpNeed policy performed well, future work may explore the use of individual student parameters to tune the model.

Central to our HelpNeed approach is our novel data-driven metric of global quality, an extension of the Hint Factory. A core insight of the work in this paper is that we can use the Hint Factory to determine step-level productivity. The original Hint Factory generated local quality values, because it assigns state values assuming the best transition between each pair of states. We developed a modified Bellman equation to determine the global quality that employs a more probabilistic approach and varying rewards for goal states based on solution optimality/length. As demonstrated in Section \ref{sec:efficiency}, the global quality successfully addresses our need for identifying step efficiency. One limitation of this work is that we consulted only one domain expert to rate a small sample of steps for assessing the ground truth of step-level productivity. As an alternative, we performed the correlation analysis in Table \ref{tab:corr} which shows that training HelpNeed defined using each of the four step efficiency metrics significantly correlates to posttest performance. In future work, we plan to further explore the proposed quality and progress metrics to better understand their tradeoffs and determine whether they can be used across domains.

\subsection{Adaptive Hint Policy}
In this section, we discuss the results of providing partially worked steps as proactive hints using our HelpNeed predictor, and our post-hoc evaluation of the predictor. Our results show that proactive hints using the predictor reduce students' Opportunistic and Far Off steps in training, and enable them to form more optimal proofs in shorter time on the posttest.

Our results showed that combining our HelpNeed predictor with proactive hints can \emph{not only} reduce the number of times students need to explicitly ask for help \emph{but also} reduce help avoidance. While one can argue that the increased number of total hints could have improved the Adaptive condition's posttest performance, the correlation analysis in section \ref{sec:hintcorr} suggests that simply receiving more proactive hints at random times can be harmful, so it is important to identify when help is needed. 

Prior research shows that gaming the system usually consists of quickly asking for successively more informative hints to get to a bottom-out hint that gives students the answer to the current question \cite{d2008developing}. Since our tutor focuses on multi-step problems where a correct answer is a set of correctly derived statements, no single hint can give a student a correct answer. We found that students in the Adaptive condition were less likely to game the system by requesting hints than the Control condition because the Adaptive group had significantly fewer steps with quick hint requests than the Control group. Another type of gaming occurs when students realize that a system predicts performance and provides proactive hints based on frequent failures \cite{d2008developing}. Since our policy uses both time and inefficient steps to determine when to proactively offer a hint, we believe it is more difficult for students to determine a pattern, and game our system. Therefore, we recommend more complex policies to make it more difficult for students to game proactive hints. Our proactive hints are more likely to occur when we predict students to derive non-optimal statements. So, even students whose goal is to game our policy, must still apply domain rules correctly, and therefore, still get valuable practice. 

We also evaluated students’ help behavior using our HelpNeed classification and predictor. We found that the Adaptive condition has significantly lower possible help avoidance and help abuse, with significantly more possible appropriate help than the Control. A limitation of this study is that this comparison relies on our HelpNeed classification and predictor, which are only shown to be correlated to posttest performance, but have not been proven to correspond directly to expert measures of help need. 

The evaluation of the HelpNeed predictor in section \ref{sec:viz} suggests ways to identify steps where students did not show effective help-seeking. Further, this evaluation helped us understand how we can improve the predictor. The HelpNeed predictor predicted that students in the Adaptive condition would carry out more steps without demonstrated HelpNeed than they were able to. We believe this resulted from not differentiating efficient steps with hints from those without during the prediction task. Beck et al. in \cite{beck2008does} proposed a Bayesian Evaluation model for a Reading Tutor to measure the impact of help on student knowledge. Further, a study by Chaudhry et al. used a deep learning model to jointly predict the hint-taking and knowledge tracing task. They found that incorporating the hint-taking model improved the deep learning knowledge tracing \cite{chaudhry2018modeling}. Similar to these studies, our predictor should be revised to account for whether or not students received and justified hints. 

\subsection{Generalization to other domains}
A limitation of this work is that we only assess our method in one ITS for one domain. To apply the HelpNeed model and predictor, the requirements for a new domain are to have state and state-transition representations (so we can define steps) and scoring for final solutions so the state-based and state-free classifiers can learn values for the HelpNeed predictor. Deriving state representations, and therefore the HelpNeed methods, should be relatively straightforward in well-structured domains such as multi-step math, physics, or statistics problems, but studies are needed to confirm their effectiveness. 

It is possible that the HelpNeed methods will also work in programming-related contexts. For example, for the iList linked list tutor, Fossati et al. \cite{fossati2015data} devised the Procedural Knowledge Model (PKM), to identify “critical” states for feedback, using per-problem models. We believe that our HelpNeed model could be used to provide a general cross-problem model, that could reduce the time needed to design per-problem models. 

Several studies have applied the Hint Factory in the domain of programming by representing problem-solving states as abstract syntax trees (ASTs) \cite{iii2014generating,rivers2013automatic,price2016generating}.   Researchers have already performed expert analyses \cite{dong2019defining} and preliminary automated methods to determine unproductivity \cite{marwan2020unproductive} in novice programming . In future work, we propose to compare the effectiveness of the HelpNeed method with these methods. We note that programming problems have large state-spaces  \cite{price2016generating,rivers2017data}, which can lead to fewer state-based matches but the methods proposed here can be used to developed state-free classifiers that do not require state matches to build a HelpNeed predictor.  

Generalization will also require adapting the definition of both Opportunistic and Far Off steps. For the specific domain and tutor in this study, we defined Opportunistic steps as inefficient steps that \textit{do not} need help, and Far Off as inefficient steps that \textit{do} need help. For other problem-solving domains, these categories should be defined based on how long a student should be allowed to work without intervention. For example, in a block-based programming problem that typically takes 1000 steps, it would be unreasonable to proactively intervene every 2 steps. We would expect to use a sliding window, tallying time and the proportion of inefficient steps, to determine when an inefficient step can be considered Opportunistic or Far Off. We would adjust the definitions of Opportunistic and Far Off steps until the definition applied to a historical dataset shows correlation with posttest performance as done here in Section \ref{sec:select_cor}. 

\section{Conclusions and Future Work}
In this paper, we present a novel approach for predicting HelpNeed in a logic tutor by extending the Hint Factory. We present a modification to the Bellman equation for quantifying the global quality of problem-solving states in well-structured open-ended domains. We show how this quality metric can be used to form a novel data-driven HelpNeed classification model for unproductivity. Overall, we found that providing partially worked steps as proactive hints upon predictions of HelpNeed can improve students’ posttest problem solving optimality and time. Furthermore, our Adaptive hint policy led students to improved training behaviors, with fewer steps predicted to need help, lower possible help avoidance and help abuse, and a higher help appropriateness (receiving help when it was predicted to be needed). This work demonstrates that our novel, data-driven HelpNeed predictor can address the assistance dilemma for well-structured open-ended problem-solving in logic. The design of our system has revealed the importance of considering pedagogy and theoretical grounding in determining HelpNeed. Specifically, we based decisions about what step-level features may be indicative of HelpNeed on prior literature on learning. Related data-driven systems should also incorporate what is known about learning with the opportunities that data-rich systems afford. 

This work has two main limitations. First, the HelpNeed predictor does not incorporate hint usage, meaning that some students with efficient steps based on hints were not given future help when it was needed. Second, the HelpNeed model and predictors here have only been evaluated in a single tutor in a single domain. While
the step efficiency and HelpNeed metrics are designed to apply to well-structured, open-ended domains that value shorter solutions, studies are needed to confirm their effectiveness for other domains.






\bibliographystyle{acmtrans}
\bibliography{./ref}

\begin{thebibliography}{}

\bibitem[\protect\citeauthoryear{Aleven and Koedinger}{Aleven and
  Koedinger}{2000}]{aleven2000limitations}
{\sc Aleven, V.} {\sc and} {\sc Koedinger, K.~R.} 2000.
\newblock Limitations of student control: Do students know when they need help?
\newblock In {\em International Conference on Intelligent Tutoring Systems}.
  Springer, 292--303.

\bibitem[\protect\citeauthoryear{Aleven, Mclaren, Roll, and Koedinger}{Aleven
  et~al\mbox{.}}{2006}]{aleven2006toward}
{\sc Aleven, V.}, {\sc Mclaren, B.}, {\sc Roll, I.}, {\sc and} {\sc Koedinger,
  K.} 2006.
\newblock Toward meta-cognitive tutoring: A model of help seeking with a
  cognitive tutor.
\newblock {\em International Journal of Artificial Intelligence in
  Education\/}~{\em 16,\/}~2, 101--128.

\bibitem[\protect\citeauthoryear{Anohina}{Anohina}{2007}]{anohina2007advances}
{\sc Anohina, A.} 2007.
\newblock Advances in intelligent tutoring systems: problem-solving modes and
  model of hints.
\newblock {\em International Journal of Computers Communications \&
  Control\/}~{\em 2,\/}~1, 48--55.

\bibitem[\protect\citeauthoryear{Arroyo, Beck, Beal, Wing, and Woolf}{Arroyo
  et~al\mbox{.}}{2001}]{arroyo2001analyzing}
{\sc Arroyo, I.}, {\sc Beck, J.~E.}, {\sc Beal, C.~R.}, {\sc Wing, R.}, {\sc
  and} {\sc Woolf, B.~P.} 2001.
\newblock Analyzing students’ response to help provision in an elementary
  mathematics intelligent tutoring system.
\newblock In {\em Papers of the AIED-2001 workshop on help provision and help
  seeking in interactive learning environments}. Citeseer, 34--46.

\bibitem[\protect\citeauthoryear{Arvai}{Arvai}{2018}]{gridsearch}
{\sc Arvai, K.} 2018.
\newblock Fine tuning a classifier in scikit-learn.
\newblock
  \url{https://towardsdatascience.com/fine-tuning-a-classifier-in-scikit-learn-66e048c21e65}.

\bibitem[\protect\citeauthoryear{Azevedo and Cromley}{Azevedo and
  Cromley}{2004}]{azevedo2004does}
{\sc Azevedo, R.} {\sc and} {\sc Cromley, J.~G.} 2004.
\newblock Does training on self-regulated learning facilitate students'
  learning with hypermedia?
\newblock {\em Journal of educational psychology\/}~{\em 96,\/}~3, 523.

\bibitem[\protect\citeauthoryear{Baker, Corbett, Koedinger, Evenson, Roll,
  Wagner, Naim, Raspat, Baker, and Beck}{Baker
  et~al\mbox{.}}{2006}]{d2006adapting}
{\sc Baker, R.~S.}, {\sc Corbett, A.~T.}, {\sc Koedinger, K.~R.}, {\sc Evenson,
  S.}, {\sc Roll, I.}, {\sc Wagner, A.~Z.}, {\sc Naim, M.}, {\sc Raspat, J.},
  {\sc Baker, D.~J.}, {\sc and} {\sc Beck, J.~E.} 2006.
\newblock Adapting to when students game an intelligent tutoring system.
\newblock In {\em International Conference on Intelligent Tutoring Systems}.
  Springer, 392--401.

\bibitem[\protect\citeauthoryear{Baker, Corbett, Roll, and Koedinger}{Baker
  et~al\mbox{.}}{2008}]{d2008developing}
{\sc Baker, R.~S.}, {\sc Corbett, A.~T.}, {\sc Roll, I.}, {\sc and} {\sc
  Koedinger, K.~R.} 2008.
\newblock Developing a generalizable detector of when students game the system.
\newblock {\em User Modeling and User-Adapted Interaction\/}~{\em 18,\/}~3,
  287--314.

\bibitem[\protect\citeauthoryear{Baker, De~Carvalho, Raspat, Aleven, Corbett,
  and Koedinger}{Baker et~al\mbox{.}}{2009}]{baker2009educational}
{\sc Baker, R.~S.}, {\sc De~Carvalho, A.}, {\sc Raspat, J.}, {\sc Aleven, V.},
  {\sc Corbett, A.~T.}, {\sc and} {\sc Koedinger, K.~R.} 2009.
\newblock Educational software features that encourage and discourage “gaming
  the system”.
\newblock In {\em Proceedings of the 14th international conference on
  artificial intelligence in education}. 475--482.

\bibitem[\protect\citeauthoryear{Barnes and Stamper}{Barnes and
  Stamper}{2010}]{barnes2010automatic}
{\sc Barnes, T.} {\sc and} {\sc Stamper, J.} 2010.
\newblock Automatic hint generation for logic proof tutoring using historical
  data.
\newblock {\em Journal of Educational Technology \& Society\/}~{\em 13,\/}~1,
  3.

\bibitem[\protect\citeauthoryear{Barnes, Stamper, and Croy}{Barnes
  et~al\mbox{.}}{2011}]{barnes2011using}
{\sc Barnes, T.}, {\sc Stamper, J.}, {\sc and} {\sc Croy, M.} 2011.
\newblock Using markov decision processes for automatic hint generation “.
\newblock {\em Handbook of Educational Data Mining\/}~{\em 467}.

\bibitem[\protect\citeauthoryear{Barnes, Stamper, Lehmann, and Croy}{Barnes
  et~al\mbox{.}}{2008}]{barnes2008pilot}
{\sc Barnes, T.}, {\sc Stamper, J.~C.}, {\sc Lehmann, L.}, {\sc and} {\sc Croy,
  M.~J.} 2008.
\newblock A pilot study on logic proof tutoring using hints generated from
  historical student data.
\newblock In {\em EDM}. 197--201.

\bibitem[\protect\citeauthoryear{Bartholom{\'e}, Stahl, Pieschl, and
  Bromme}{Bartholom{\'e} et~al\mbox{.}}{2006}]{bartholome2006matters}
{\sc Bartholom{\'e}, T.}, {\sc Stahl, E.}, {\sc Pieschl, S.}, {\sc and} {\sc
  Bromme, R.} 2006.
\newblock What matters in help-seeking? a study of help effectiveness and
  learner-related factors.
\newblock {\em Computers in Human Behavior\/}~{\em 22,\/}~1, 113--129.

\bibitem[\protect\citeauthoryear{Beck, Chang, Mostow, and Corbett}{Beck
  et~al\mbox{.}}{2008}]{beck2008does}
{\sc Beck, J.~E.}, {\sc Chang, K.-m.}, {\sc Mostow, J.}, {\sc and} {\sc
  Corbett, A.} 2008.
\newblock Does help help? introducing the bayesian evaluation and assessment
  methodology.
\newblock In {\em International Conference on Intelligent Tutoring Systems}.
  Springer, 383--394.

\bibitem[\protect\citeauthoryear{Beck and Gong}{Beck and
  Gong}{2013}]{beck2013wheel}
{\sc Beck, J.~E.} {\sc and} {\sc Gong, Y.} 2013.
\newblock Wheel-spinning: Students who fail to master a skill.
\newblock In {\em International conference on artificial intelligence in
  education}. Springer, 431--440.

\bibitem[\protect\citeauthoryear{Borek, McLaren, Karabinos, and Yaron}{Borek
  et~al\mbox{.}}{2009}]{borek2009much}
{\sc Borek, A.}, {\sc McLaren, B.~M.}, {\sc Karabinos, M.}, {\sc and} {\sc
  Yaron, D.} 2009.
\newblock How much assistance is helpful to students in discovery learning?
\newblock In {\em European Conference on Technology Enhanced Learning}.
  Springer, 391--404.

\bibitem[\protect\citeauthoryear{Botelho, Varatharaj, Patikorn, Doherty, Adjei,
  and Beck}{Botelho et~al\mbox{.}}{2019}]{botelho2019developing}
{\sc Botelho, A.}, {\sc Varatharaj, A.}, {\sc Patikorn, T.}, {\sc Doherty, D.},
  {\sc Adjei, S.}, {\sc and} {\sc Beck, J.} 2019.
\newblock Developing early detectors of student attrition and wheel spinning
  using deep learning.
\newblock {\em IEEE Transactions on Learning Technologies\/}.

\bibitem[\protect\citeauthoryear{Bunt and Conati}{Bunt and
  Conati}{2003}]{bunt2003probabilistic}
{\sc Bunt, A.} {\sc and} {\sc Conati, C.} 2003.
\newblock Probabilistic student modelling to improve exploratory behaviour.
\newblock {\em User Modeling and User-Adapted Interaction\/}~{\em 13,\/}~3,
  269--309.

\bibitem[\protect\citeauthoryear{Bunt, Conati, and Muldner}{Bunt
  et~al\mbox{.}}{2004}]{bunt2004scaffolding}
{\sc Bunt, A.}, {\sc Conati, C.}, {\sc and} {\sc Muldner, K.} 2004.
\newblock Scaffolding self-explanation to improve learning in exploratory
  learning environments.
\newblock In {\em International Conference on Intelligent Tutoring Systems}.
  Springer, 656--667.

\bibitem[\protect\citeauthoryear{Capraro, An, Ma, Rangel-Chavez, and
  Harbaugh}{Capraro et~al\mbox{.}}{2012}]{capraro2012investigation}
{\sc Capraro, M.~M.}, {\sc An, S.~A.}, {\sc Ma, T.}, {\sc Rangel-Chavez,
  A.~F.}, {\sc and} {\sc Harbaugh, A.} 2012.
\newblock An investigation of preservice teachers’ use of guess and check in
  solving a semi open-ended mathematics problem.
\newblock {\em The Journal of Mathematical Behavior\/}~{\em 31,\/}~1, 105--116.

\bibitem[\protect\citeauthoryear{Cen, Koedinger, and Junker}{Cen
  et~al\mbox{.}}{2007}]{cen2007over}
{\sc Cen, H.}, {\sc Koedinger, K.~R.}, {\sc and} {\sc Junker, B.} 2007.
\newblock Is over practice necessary?-improving learning efficiency with the
  cognitive tutor through educational data mining.
\newblock {\em Frontiers in artificial intelligence and applications\/}~{\em
  158}, 511.

\bibitem[\protect\citeauthoryear{Chaudhry, Singh, Dogga, and Saini}{Chaudhry
  et~al\mbox{.}}{2018}]{chaudhry2018modeling}
{\sc Chaudhry, R.}, {\sc Singh, H.}, {\sc Dogga, P.}, {\sc and} {\sc Saini,
  S.~K.} 2018.
\newblock Modeling hint-taking behavior and knowledge state of students with
  multi-task learning.
\newblock {\em International Educational Data Mining Society\/}.

\bibitem[\protect\citeauthoryear{Chen and Guestrin}{Chen and
  Guestrin}{2016}]{chen2016xgboost}
{\sc Chen, T.} {\sc and} {\sc Guestrin, C.} 2016.
\newblock Xgboost: A scalable tree boosting system.
\newblock In {\em Proceedings of the 22nd acm sigkdd international conference
  on knowledge discovery and data mining}. ACM, 785--794.

\bibitem[\protect\citeauthoryear{Cody and Mostafavi}{Cody and
  Mostafavi}{2017}]{cody2017investigating}
{\sc Cody, C.} {\sc and} {\sc Mostafavi, B.} 2017.
\newblock Investigating the impact of unsolicited next-step and subgoal hints
  on dropout in a logic proof tutor.
\newblock In {\em Proceedings of the 2017 ACM SIGCSE Technical Symposium on
  Computer Science Education}. ACM, 705--705.

\bibitem[\protect\citeauthoryear{Conati, Gertner, and Vanlehn}{Conati
  et~al\mbox{.}}{2002}]{conati2002using}
{\sc Conati, C.}, {\sc Gertner, A.}, {\sc and} {\sc Vanlehn, K.} 2002.
\newblock Using bayesian networks to manage uncertainty in student modeling.
\newblock {\em User modeling and user-adapted interaction\/}~{\em 12,\/}~4,
  371--417.

\bibitem[\protect\citeauthoryear{Corbett and Anderson}{Corbett and
  Anderson}{1994}]{corbett1994knowledge}
{\sc Corbett, A.~T.} {\sc and} {\sc Anderson, J.~R.} 1994.
\newblock Knowledge tracing: Modeling the acquisition of procedural knowledge.
\newblock {\em User modeling and user-adapted interaction\/}~{\em 4,\/}~4,
  253--278.

\bibitem[\protect\citeauthoryear{Dong, Marwan, Catete, Price, and Barnes}{Dong
  et~al\mbox{.}}{2019}]{dong2019defining}
{\sc Dong, Y.}, {\sc Marwan, S.}, {\sc Catete, V.}, {\sc Price, T.}, {\sc and}
  {\sc Barnes, T.} 2019.
\newblock Defining tinkering behavior in open-ended block-based programming
  assignments.
\newblock In {\em Proceedings of the 50th ACM Technical Symposium on Computer
  Science Education}. 1204--1210.

\bibitem[\protect\citeauthoryear{Eagle, Hicks, and Barnes}{Eagle
  et~al\mbox{.}}{2015}]{eagle2015interaction}
{\sc Eagle, M.}, {\sc Hicks, D.}, {\sc and} {\sc Barnes, T.} 2015.
\newblock Interaction network estimation: Predicting problem-solving diversity
  in interactive environments.
\newblock {\em International Educational Data Mining Society\/}.

\bibitem[\protect\citeauthoryear{Fossati, Di~Eugenio, Ohlsson, Brown, and
  Chen}{Fossati et~al\mbox{.}}{2015}]{fossati2015data}
{\sc Fossati, D.}, {\sc Di~Eugenio, B.}, {\sc Ohlsson, S.}, {\sc Brown, C.},
  {\sc and} {\sc Chen, L.} 2015.
\newblock Data driven automatic feedback generation in the ilist intelligent
  tutoring system.
\newblock {\em Technology, Instruction, Cognition and Learning\/}~{\em
  10,\/}~1, 5--26.

\bibitem[\protect\citeauthoryear{Fratamico, Conati, Kardan, and Roll}{Fratamico
  et~al\mbox{.}}{2017}]{fratamico2017applying}
{\sc Fratamico, L.}, {\sc Conati, C.}, {\sc Kardan, S.}, {\sc and} {\sc Roll,
  I.} 2017.
\newblock Applying a framework for student modeling in exploratory learning
  environments: Comparing data representation granularity to handle environment
  complexity.
\newblock {\em International Journal of Artificial Intelligence in
  Education\/}~{\em 27,\/}~2, 320--352.

\bibitem[\protect\citeauthoryear{Iii, Hicks, and Barnes}{Iii
  et~al\mbox{.}}{2014}]{iii2014generating}
{\sc Iii, B.~P.}, {\sc Hicks, A.}, {\sc and} {\sc Barnes, T.} 2014.
\newblock Generating hints for programming problems using intermediate output.
\newblock In {\em Educational Data Mining 2014}. Citeseer.

\bibitem[\protect\citeauthoryear{Kai, Almeda, Baker, Heffernan, and
  Heffernan}{Kai et~al\mbox{.}}{2018}]{kai2018decision}
{\sc Kai, S.}, {\sc Almeda, M.~V.}, {\sc Baker, R.~S.}, {\sc Heffernan, C.},
  {\sc and} {\sc Heffernan, N.} 2018.
\newblock Decision tree modeling of wheel-spinning and productive persistence
  in skill builders.
\newblock {\em JEDM| Journal of Educational Data Mining\/}~{\em 10,\/}~1,
  36--71.

\bibitem[\protect\citeauthoryear{Kardan and Conati}{Kardan and
  Conati}{2015}]{kardan2015providing}
{\sc Kardan, S.} {\sc and} {\sc Conati, C.} 2015.
\newblock Providing adaptive support in an interactive simulation for learning:
  An experimental evaluation.
\newblock In {\em Proceedings of the 33rd Annual ACM Conference on Human
  Factors in Computing Systems}. ACM, 3671--3680.

\bibitem[\protect\citeauthoryear{Kinnebrew, Segedy, and Biswas}{Kinnebrew
  et~al\mbox{.}}{2014}]{kinnebrew2014analyzing}
{\sc Kinnebrew, J.~S.}, {\sc Segedy, J.~R.}, {\sc and} {\sc Biswas, G.} 2014.
\newblock Analyzing the temporal evolution of students’ behaviors in
  open-ended learning environments.
\newblock {\em Metacognition and learning\/}~{\em 9,\/}~2, 187--215.

\bibitem[\protect\citeauthoryear{Klahr}{Klahr}{2009}]{klahr2009every}
{\sc Klahr, D.} 2009.
\newblock " to every thing there is a season, and a time to every purpose under
  the heavens": What about direct instruction?

\bibitem[\protect\citeauthoryear{Kock and Paramythis}{Kock and
  Paramythis}{2010}]{kock2010towards}
{\sc Kock, M.} {\sc and} {\sc Paramythis, A.} 2010.
\newblock Towards adaptive learning support on the basis of behavioural
  patterns in learning activity sequences.
\newblock In {\em 2010 International Conference on Intelligent Networking and
  Collaborative Systems}. IEEE, 100--107.

\bibitem[\protect\citeauthoryear{Koedinger and Aleven}{Koedinger and
  Aleven}{2007}]{koedinger2007exploring}
{\sc Koedinger, K.~R.} {\sc and} {\sc Aleven, V.} 2007.
\newblock Exploring the assistance dilemma in experiments with cognitive
  tutors.
\newblock {\em Educational Psychology Review\/}~{\em 19,\/}~3, 239--264.

\bibitem[\protect\citeauthoryear{Maniktala, Cody, Barnes, and Chi}{Maniktala
  et~al\mbox{.}}{2020}]{maniktala2020leveraging}
{\sc Maniktala, M.}, {\sc Cody, C.}, {\sc Barnes, T.}, {\sc and} {\sc Chi, M.}
  2020.
\newblock Avoiding help avoidance: Using interface design changes to promote
  unsolicited hint usage in an intelligent tutor.
\newblock {\em Accepted to appear in: International Journal of Artificial
  Intelligence in Education\/}.

\bibitem[\protect\citeauthoryear{Marwan, Dombe, and Price}{Marwan
  et~al\mbox{.}}{2020}]{marwan2020unproductive}
{\sc Marwan, S.}, {\sc Dombe, A.}, {\sc and} {\sc Price, T.~W.} 2020.
\newblock Unproductive help-seeking in programming: What it is and how to
  address it.
\newblock In {\em Proceedings of the 2020 ACM Conference on Innovation and
  Technology in Computer Science Education}. 54--60.

\bibitem[\protect\citeauthoryear{Mayer}{Mayer}{1992}]{mayer1992thinking}
{\sc Mayer, R.~E.} 1992.
\newblock {\em Thinking, problem solving, cognition}.
\newblock WH Freeman/Times Books/Henry Holt \& Co.

\bibitem[\protect\citeauthoryear{McLaren, Lim, and Koedinger}{McLaren
  et~al\mbox{.}}{2008}]{mclaren2008and}
{\sc McLaren, B.~M.}, {\sc Lim, S.-J.}, {\sc and} {\sc Koedinger, K.~R.} 2008.
\newblock When and how often should worked examples be given to students? new
  results and a summary of the current state of research.
\newblock In {\em Proceedings of the 30th annual conference of the cognitive
  science society}. 2176--2181.

\bibitem[\protect\citeauthoryear{McLaren, Timms, Weihnacht, Brenner, Luttgen,
  Grillo-Hill, and Brown}{McLaren et~al\mbox{.}}{2014}]{mclaren2014web}
{\sc McLaren, B.~M.}, {\sc Timms, M.}, {\sc Weihnacht, D.}, {\sc Brenner, D.},
  {\sc Luttgen, K.}, {\sc Grillo-Hill, A.}, {\sc and} {\sc Brown, D.~H.} 2014.
\newblock A web-based system to support inquiry learning.
\newblock In {\em Proceedings of the 6th International Conference on Computer
  Supported Education-Volume 1}. SCITEPRESS-Science and Technology
  Publications, Lda, 43--52.

\bibitem[\protect\citeauthoryear{Merceron and Yacef}{Merceron and
  Yacef}{2005}]{merceron2005educational}
{\sc Merceron, A.} {\sc and} {\sc Yacef, K.} 2005.
\newblock Educational data mining: a case study.
\newblock In {\em AIED}. 467--474.

\bibitem[\protect\citeauthoryear{Mostafavi and Barnes}{Mostafavi and
  Barnes}{2017}]{mostafavi2017evolution}
{\sc Mostafavi, B.} {\sc and} {\sc Barnes, T.} 2017.
\newblock Evolution of an intelligent deductive logic tutor using data-driven
  elements.
\newblock {\em International Journal of Artificial Intelligence in
  Education\/}~{\em 27,\/}~1, 5--36.

\bibitem[\protect\citeauthoryear{Mostafavi, Liu, and Barnes}{Mostafavi
  et~al\mbox{.}}{2015}]{mostafavi2015data}
{\sc Mostafavi, B.}, {\sc Liu, Z.}, {\sc and} {\sc Barnes, T.} 2015.
\newblock Data-driven proficiency profiling.
\newblock {\em International Educational Data Mining Society\/}.

\bibitem[\protect\citeauthoryear{Murray and VanLehn}{Murray and
  VanLehn}{2005}]{murray2005effects}
{\sc Murray, R.~C.} {\sc and} {\sc VanLehn, K.} 2005.
\newblock Effects of dissuading unnecessary help requests while providing
  proactive help.
\newblock In {\em AIED}. Citeseer, 887--889.

\bibitem[\protect\citeauthoryear{Murray and VanLehn}{Murray and
  VanLehn}{2006}]{murray2006comparison}
{\sc Murray, R.~C.} {\sc and} {\sc VanLehn, K.} 2006.
\newblock A comparison of decision-theoretic, fixed-policy and random tutorial
  action selection.
\newblock In {\em International Conference on Intelligent Tutoring Systems}.
  Springer, 114--123.

\bibitem[\protect\citeauthoryear{Murray, Vanlehn, and Mostow}{Murray
  et~al\mbox{.}}{2004}]{murray2004looking}
{\sc Murray, R.~C.}, {\sc Vanlehn, K.}, {\sc and} {\sc Mostow, J.} 2004.
\newblock Looking ahead to select tutorial actions: A decision-theoretic
  approach.
\newblock {\em International Journal of Artificial Intelligence in
  Education\/}~{\em 14,\/}~3, 4, 235--278.

\bibitem[\protect\citeauthoryear{Park and Matsuda}{Park and
  Matsuda}{2018}]{park2018predicting}
{\sc Park, S.} {\sc and} {\sc Matsuda, N.} 2018.
\newblock Predicting students’ unproductive failure on intelligent tutors in
  adaptive online courseware.
\newblock In {\em Proceedings of the Sixth Annual GIFT Users Symposium}.
  Vol.~6. US Army Research Laboratory, 131.

\bibitem[\protect\citeauthoryear{Pedregosa, Varoquaux, Gramfort, Michel,
  Thirion, Grisel, Blondel, Prettenhofer, Weiss, Dubourg,
  et~al\mbox{.}}{Pedregosa et~al\mbox{.}}{2011}]{pedregosa2011scikit}
{\sc Pedregosa, F.}, {\sc Varoquaux, G.}, {\sc Gramfort, A.}, {\sc Michel, V.},
  {\sc Thirion, B.}, {\sc Grisel, O.}, {\sc Blondel, M.}, {\sc Prettenhofer,
  P.}, {\sc Weiss, R.}, {\sc Dubourg, V.}, {\sc et~al\mbox{.}} 2011.
\newblock Scikit-learn: Machine learning in python.
\newblock {\em Journal of machine learning research\/}~{\em 12,\/}~Oct,
  2825--2830.

\bibitem[\protect\citeauthoryear{Pe{\~n}a, Kayashima, Mizoguchi, and
  Dominguez}{Pe{\~n}a et~al\mbox{.}}{2011}]{pena2011improving}
{\sc Pe{\~n}a, A.}, {\sc Kayashima, M.}, {\sc Mizoguchi, R.}, {\sc and} {\sc
  Dominguez, R.} 2011.
\newblock Improving students’ meta-cognitive skills within intelligent
  educational systems: A review.
\newblock In {\em International Conference on Foundations of Augmented
  Cognition}. Springer, 442--451.

\bibitem[\protect\citeauthoryear{Polya}{Polya}{2004}]{polya2004solve}
{\sc Polya, G.} 2004.
\newblock {\em How to solve it: A new aspect of mathematical method}. Vol.~85.
\newblock Princeton university press.

\bibitem[\protect\citeauthoryear{Price, Dong, and Barnes}{Price
  et~al\mbox{.}}{2016}]{price2016generating}
{\sc Price, T.~W.}, {\sc Dong, Y.}, {\sc and} {\sc Barnes, T.} 2016.
\newblock Generating data-driven hints for open-ended programming.
\newblock {\em International Educational Data Mining Society\/}.

\bibitem[\protect\citeauthoryear{Price, Liu, Catet{\'e}, and Barnes}{Price
  et~al\mbox{.}}{2017}]{price2017factors}
{\sc Price, T.~W.}, {\sc Liu, Z.}, {\sc Catet{\'e}, V.}, {\sc and} {\sc Barnes,
  T.} 2017.
\newblock Factors influencing students' help-seeking behavior while programming
  with human and computer tutors.
\newblock In {\em Proceedings of the 2017 ACM Conference on International
  Computing Education Research}. ACM, 127--135.

\bibitem[\protect\citeauthoryear{Price, Zhi, and Barnes}{Price
  et~al\mbox{.}}{2017}]{price2017hint}
{\sc Price, T.~W.}, {\sc Zhi, R.}, {\sc and} {\sc Barnes, T.} 2017.
\newblock Hint generation under uncertainty: The effect of hint quality on
  help-seeking behavior.
\newblock In {\em International Conference on Artificial Intelligence in
  Education}. Springer, 311--322.

\bibitem[\protect\citeauthoryear{Puustinen}{Puustinen}{1998}]{puustinen1998help}
{\sc Puustinen, M.} 1998.
\newblock Help-seeking behavior in a problem-solving situation: Development of
  self-regulation.
\newblock {\em European Journal of Psychology of education\/}~{\em 13,\/}~2,
  271.

\bibitem[\protect\citeauthoryear{Rivers and Koedinger}{Rivers and
  Koedinger}{2013}]{rivers2013automatic}
{\sc Rivers, K.} {\sc and} {\sc Koedinger, K.~R.} 2013.
\newblock Automatic generation of programming feedback: A data-driven approach.
\newblock In {\em The First Workshop on AI-supported Education for Computer
  Science (AIEDCS 2013)}. Vol.~50.

\bibitem[\protect\citeauthoryear{Rivers and Koedinger}{Rivers and
  Koedinger}{2017}]{rivers2017data}
{\sc Rivers, K.} {\sc and} {\sc Koedinger, K.~R.} 2017.
\newblock Data-driven hint generation in vast solution spaces: a self-improving
  python programming tutor.
\newblock {\em International Journal of Artificial Intelligence in
  Education\/}~{\em 27,\/}~1, 37--64.

\bibitem[\protect\citeauthoryear{Roll, Aleven, McLaren, and Koedinger}{Roll
  et~al\mbox{.}}{2011}]{roll2011improving}
{\sc Roll, I.}, {\sc Aleven, V.}, {\sc McLaren, B.~M.}, {\sc and} {\sc
  Koedinger, K.~R.} 2011.
\newblock Improving students’ help-seeking skills using metacognitive
  feedback in an intelligent tutoring system.
\newblock {\em Learning and instruction\/}~{\em 21,\/}~2, 267--280.

\bibitem[\protect\citeauthoryear{Rus, Banjade, Niraula, Gire, and
  Franceschetti}{Rus et~al\mbox{.}}{2017}]{rus2017study}
{\sc Rus, V.}, {\sc Banjade, R.}, {\sc Niraula, N.}, {\sc Gire, E.}, {\sc and}
  {\sc Franceschetti, D.} 2017.
\newblock A study on two hint-level policies in conversational intelligent
  tutoring systems.
\newblock In {\em Innovations in Smart Learning}. Springer, 175--184.

\bibitem[\protect\citeauthoryear{Smith}{Smith}{2012}]{smith2012toward}
{\sc Smith, M.~U.} 2012.
\newblock {\em Toward a unified theory of problem solving: Views from the
  content domains}.
\newblock Routledge.

\bibitem[\protect\citeauthoryear{Stamper and Barnes}{Stamper and
  Barnes}{2009}]{stamper2009unsupervised}
{\sc Stamper, J.} {\sc and} {\sc Barnes, T.} 2009.
\newblock Unsupervised mdp value selection for automating its capabilities.
\newblock {\em International Working Group on Educational Data Mining\/}.

\bibitem[\protect\citeauthoryear{Stamper, Barnes, Lehmann, and Croy}{Stamper
  et~al\mbox{.}}{2008}]{stamper2008hint}
{\sc Stamper, J.}, {\sc Barnes, T.}, {\sc Lehmann, L.}, {\sc and} {\sc Croy,
  M.} 2008.
\newblock The hint factory: Automatic generation of contextualized help for
  existing computer aided instruction.
\newblock In {\em Proceedings of the 9th International Conference on
  Intelligent Tutoring Systems Young Researchers Track}. 71--78.

\bibitem[\protect\citeauthoryear{Tch{\'e}tagni and Nkambou}{Tch{\'e}tagni and
  Nkambou}{2002}]{tchetagni2002hierarchical}
{\sc Tch{\'e}tagni, J.~M.} {\sc and} {\sc Nkambou, R.} 2002.
\newblock Hierarchical representation and evaluation of the student in an
  intelligent tutoring system.
\newblock In {\em International Conference on Intelligent Tutoring Systems}.
  Springer, 708--717.

\bibitem[\protect\citeauthoryear{Ueno and Miyazawa}{Ueno and
  Miyazawa}{2017}]{ueno2017irt}
{\sc Ueno, M.} {\sc and} {\sc Miyazawa, Y.} 2017.
\newblock Irt-based adaptive hints to scaffold learning in programming.
\newblock {\em IEEE Transactions on Learning Technologies\/}~{\em 11,\/}~4,
  415--428.

\bibitem[\protect\citeauthoryear{VanLehn}{VanLehn}{2011}]{vanlehn2011relative}
{\sc VanLehn, K.} 2011.
\newblock The relative effectiveness of human tutoring, intelligent tutoring
  systems, and other tutoring systems.
\newblock {\em Educational Psychologist\/}~{\em 46,\/}~4, 197--221.

\bibitem[\protect\citeauthoryear{Wood and Wood}{Wood and
  Wood}{1999}]{wood1999help}
{\sc Wood, H.} {\sc and} {\sc Wood, D.} 1999.
\newblock Help seeking, learning and contingent tutoring.
\newblock {\em Computers \& Education\/}~{\em 33,\/}~2-3, 153--169.

\bibitem[\protect\citeauthoryear{Yacef}{Yacef}{2005}]{yacef2005logic}
{\sc Yacef, K.} 2005.
\newblock The logic-ita in the classroom: a medium scale experiment.
\newblock {\em International Journal of Artificial Intelligence in
  Education\/}~{\em 15,\/}~1, 41--62.

\bibitem[\protect\citeauthoryear{Zhou, Azizsoltani, Ausin, Barnes, and
  Chi}{Zhou et~al\mbox{.}}{2019}]{zhou2019hierarchical}
{\sc Zhou, G.}, {\sc Azizsoltani, H.}, {\sc Ausin, M.~S.}, {\sc Barnes, T.},
  {\sc and} {\sc Chi, M.} 2019.
\newblock Hierarchical reinforcement learning for pedagogical policy induction.
\newblock In {\em International Conference on Artificial Intelligence in
  Education}. Springer, 544--556.

\end{thebibliography}

\clearpage
\appendix

\section{APPENDIX: Proof of Convergence for the Modified Bellman Backup function}
\label{aconv}

\textbf{Theorem}: The modified Value iteration (Eqn \ref{eq:global_quality}) converges to $GQV^{*}$ for any initial estimate $GQV$, i.e., 

\begin{equation}
\notag
    \lim_{k \rightarrow \infty} GQV_{k} = GQV^* \quad \forall  GQV
\end{equation}

For any estimate of the value function $GQV$, we define the modified Bellman backup operator $\hat{B} : R^{|S|} \to R^{|S|}$
\begin{equation}
\notag
    \hat{B}GQV(s) = GR(s) + \gamma \sum_{s'\in S}P(s'|s)GQV(s')
\end{equation}

Before we provide the proof of convergence, we provide the proof of contraction, i.e, for any two value functions GQV and GQV':
\begin{equation}
\notag
||\hat{B}GQV_k - \hat{B}GQV_k'|| \leq \gamma ||GQV_k - GQV_k'||
\end{equation}

where the max norm: 
\begin{equation}
\notag
||GQV|| = \max_{s\in S}|GQV(s)|
\end{equation}
$||v - v'||$ = Infinity norm (max difference over all states)

\textbf{Proof of contraction}:
\begin{equation}
\notag
\notag
    ||\hat{B}GQV - \hat{B}GQV'||
\end{equation}

\begin{equation}
\notag
= \left| \left| \left[GR(s) + \gamma \sum_{s'\in S}P(s'|s)GQV(s') \right] - \left[ GR(s) + \gamma \sum_{s'\in S}P(s'|s)GQV'(s')\right] \right|\right|
\end{equation}

\begin{equation}
\notag
= \gamma \left| \left| \left[ \sum_{s'\in S}P(s'|s)GQV(s') - \sum_{s'\in S}P(s'|s)GQV'(s')\right] \right|\right|
\end{equation}

\begin{equation}
\notag
= \gamma \left| \left| \left[ \sum_{s'\in S}P(s'|s)(GQV(s') - GQV'(s'))\right] \right|\right|
\end{equation}

\begin{equation}
\notag
\leq \gamma \max_s \sum_{s'\in S}P(s'|s)|GQV(s') - GQV'(s')|
\end{equation}

\begin{equation}
\notag
\leq \gamma \sum_{s'\in S}P(s'|s)||GQV - GQV'||
\end{equation}

\begin{equation}
\notag
= \gamma ||GQV - GQV'||
\end{equation}

since $P(s'|s)$ are non-negative and sum to one

\textbf{Proof of Convergence}:
\begin{equation}
\notag
\left| \left| GQV_{k+1} - GQV^*\right| \right|_\infty = \left| \left| \hat{B}GQV_{k} - GQV^*\right| \right|_\infty \leq \gamma \left| \left| GQV_{k} =GQ V^*\right|  \right|_\infty\leq ...
\end{equation}

\begin{equation}
\notag
\leq \gamma^{k+1}\left| \left| GQV_0 - GQV^*\right| \right|_\infty \longrightarrow  0
\end{equation}

\FloatBarrier 
\clearpage

\section{List of all the Features Engineered along with the level at which they were aggregated (Step (s), Problem (p), Total (t))} \label{aa}

\begin{longtable}{|l|l|l|}
\hline
Feature & Level & Description \\ \hline
GAP & s & Absolute progress on the current state using global quality \\ \hline
GRP & s & Relative progress of the current state using global quality \\ \hline
LAP & s & Absolute progress of the current state using local quality \\ \hline
LRP & s & Relative progress of the current state using local quality \\ \hline
localPrevious & s & Local Quality of the previous state \\ \hline
globalPrevious & s & Global Quality of the previous state\\ \hline
localCurrent & s & Local Quality of the current state \\ \hline
globalCurrent & s & Global Quality of the current state \\ \hline
Time & s, p, t & Time Taken \\ \hline
AvgStepTime & p, t & Average Step Time \\ \hline
ActionCount & s, p, t & \begin{tabular}[c]{@{}l@{}}Total number of actions performed (e.g. selecting nodes,\\ rule clicks, hint button clicks, etc).\end{tabular} \\ \hline
\begin{tabular}[c]{@{}l@{}}DirectProof\\ ActionCount\end{tabular} & s, p, t & \begin{tabular}[c]{@{}l@{}}The number of actions performed while working on the\\ problem as a direct proof. Students can switch between\\  direct and indirect proof at any given time.\end{tabular} \\ \hline
\begin{tabular}[c]{@{}l@{}}IndirectProof\\ ActionCount\end{tabular} & s, p, t & \begin{tabular}[c]{@{}l@{}}The number of actions performed while working on the\\ problem as an indirect proof. A logic proof is said to be\\ indirect if we first assume the negation of conclusion to\\ be true and then arrive at a contradiction.\end{tabular} \\ \hline
DirectionChange & s, p, t & \begin{tabular}[c]{@{}l@{}}The number of time a student switched between direct\\ and indirect proof using the direction switch button.\end{tabular} \\ \hline
FDActionCount & s, p, t & \begin{tabular}[c]{@{}l@{}}The number of actions performed while working on a\\ forward step. The student can switch between forward\\  and backward step at any time\end{tabular} \\ \hline
BDActionCount & s, p, t & \begin{tabular}[c]{@{}l@{}}The number of actions performed while working on a\\ step backwards using instantiation and generalization\\ logic rules.\end{tabular} \\ \hline
StepCount & p, t & \begin{tabular}[c]{@{}l@{}}The number of steps performed\end{tabular} \\ \hline
SolSize & p & \begin{tabular}[c]{@{}l@{}} The solution size of a problem, calculated as the number of the \\logic statements in the current state\end{tabular} \\ \hline
RuleDescription & s, p, t & \begin{tabular}[c]{@{}l@{}}The number of times a student clicked the description\\ button of any logic rule.\end{tabular} \\ \hline
HintRequest & s, p, t & \begin{tabular}[c]{@{}l@{}}The number of time the hint was requested by the\\ student using the “Get Suggestion” button\end{tabular} \\ \hline
ProactiveHintCount & s, p, t & The number of proactive hints given. \\ \hline
OnDemandHintCount & s, p, t & The number of on-demand hints given. \\ \hline
Deleted & p, t & The number of nodes deleted. \\ \hline
RightApp & p, t & The number of Right logic rule applications. \\ \hline
WrongApp & s, p, t & The number of Wrong logic rule applications. \\ \hline
Accuracy & s, p, t & Proportion of total logic rule applications that are correct. \\ \hline
SessionCount & t & \begin{tabular}[c]{@{}l@{}}Total number of sessions - each time a student logs back\\ in, we increment the number of sessions.\end{tabular} \\ \hline
NewSession & p & Binary variable: 1 for the first problem in any session \\ \hline
Skips & t & Total number of problems skipped. \\ \hline
Restarts & t & Total number of problems restarted. \\ \hline
EasyProblems & t & The total number of easy problems solved. \\ \hline
DifficultProblems & t & The number of difficulty problems solved. \\ \hline
\label{tab:my-table}
\end{longtable}

\FloatBarrier 
\clearpage
\section{Descriptive Statistics for the Features Selected for State-Based and State-Free Random Forest Classifier with Expert Weights (Grey cells suggest inapplicable)} \label{ab}
\FloatBarrier

\begin{longtable}{|l|l|l|l|l|}

\hline
\multicolumn{1}{|c|}{Feature} & \multicolumn{1}{c|}{State-Based} & \multicolumn{1}{c|}{State-Free} & \multicolumn{1}{c|}{Mean} & \multicolumn{1}{c|}{SD} \\ \hline
GAP                & Yes                              & \cellcolor[HTML]{656565}        & 9.91                      & 32.77                   \\ \hline
GRP                & Yes                              & \cellcolor[HTML]{656565}        & 4.26                      & 21.27                   \\ \hline
LAP                & Yes                              & \cellcolor[HTML]{656565}        & 33.85                     & 37.70                   \\ \hline
LRP                & Yes                              & \cellcolor[HTML]{656565}        & 9.05                      & 20.90                   \\ \hline
localPrevious               & Yes                              & \cellcolor[HTML]{656565}        & 31.22                     & 33.20                   \\ \hline
globalPrevious              & Yes                              & \cellcolor[HTML]{656565}        & 50.68                     & 32.06                   \\ \hline
localCurrent              & Yes                              & \cellcolor[HTML]{656565}        & 40.26                     & 37.76                   \\ \hline
globalCurrent             & Yes                              & \cellcolor[HTML]{656565}        & 54.94                     & 34.31                   \\ \hline
pTime                         & Yes                              & Yes                             & 718.55                    & 4051.65                 \\ \hline
pAvgstepTimePS                & Yes                              & Yes                             & 130.39                    & 1306.00                 \\ \hline
tAvgstepTimePS                & Yes                              & Yes                             & 266.36                    & 429.40                  \\ \hline
sActionCount                  & No                               & Yes                             & 5.98                      & 6.47                    \\ \hline
pActionCount                  & Yes                              & Yes                             & 58.60                     & 98.97                   \\ \hline
tActionCount                  & Yes                              & Yes                             & 1304.02                   & 870.32                  \\ \hline
pDirectProofActionCount       & Yes                              & Yes                             & 54.06                     & 89.52                   \\ \hline
tDirectProofActionCount       & Yes                              & Yes                             & 1204.48                   & 795.68                  \\ \hline
pInDirectProofActionCount     & No                               & Yes                             & 4.55                      & 33.72                   \\ \hline
tInDirectProofActionCount     & No                               & Yes                             & 99.54                     & 216.65                  \\ \hline
pDirectionChanges             & Yes                              & No                              & 0.17                      & 0.97                    \\ \hline
pFDActionCount                & Yes                              & Yes                             & 58.03                     & 97.75                   \\ \hline
tFDActionCount                & Yes                              & Yes                             & 1283.66                   & 855.38                  \\ \hline
pBDActionCount                & Yes                              & Yes                             & 0.58                      & 2.97                    \\ \hline
tBDActionCount                & Yes                              & Yes                             & 20.36                     & 33.04                   \\ \hline
tStepCount                    & No                               & Yes                             & 119.40                    & 76.72                   \\ \hline
pSolSize                      & Yes                              & Yes                             & 7.45                      & 3.00                    \\ \hline
sRuleDescription              & Yes                              & No                              & 0.57                      & 2.34                    \\ \hline
pRuleDescription              & Yes                              & Yes                             & 7.46                      & 25.68                   \\ \hline
tRuleDescription              & Yes                              & Yes                             & 196.96                    & 230.58                  \\ \hline
pHintRequest                  & Yes                              & Yes                             & 1.64                      & 5.48                    \\ \hline
tHintRequest                  & Yes                              & Yes                             & 11.60                     & 19.83                   \\ \hline
sDeleted                      & No                               & Yes                             & 0.13                      & 0.36                    \\ \hline
pDeleted                      & Yes                              & Yes                             & 1.71                      & 4.53                    \\ \hline
tDeleted                      & Yes                              & Yes                             & 28.95                     & 27.32                   \\ \hline
pRightApp                     & Yes                              & Yes                             & 6.42                      & 7.10                    \\ \hline
tRightApp                     & Yes                              & Yes                             & 119.40                    & 76.72                   \\ \hline
sWrongApp                     & Yes                              & Yes                             & 0.45                      & 1.24                    \\ \hline
pWrongApp                     & Yes                              & Yes                             & 4.92                      & 10.51                   \\ \hline
tWrongApp                     & Yes                              & Yes                             & 246.16                    & 162.78                  \\ \hline
sAccuracy                     & No                               & Yes                             & 0.69                      & 0.41                    \\ \hline
pAccuracy                     & No                               & Yes                             & 0.73                      & 0.25                    \\ \hline
tSessionCount                 & Yes                              & No                              & 3.11                      & 2.22                    \\ \hline
pNewSession                   & Yes                              & No                              & 0.05                      & 0.21                    \\ \hline

pRestarts                     & Yes                              & Yes                             & 0.13                      & 0.52                    \\ \hline
tRestarts                     & No                               & Yes                             & 4.79                      & 5.42                    \\ \hline
tEasyProblemsPS               & Yes                              & Yes                             & 3.59                      & 2.75                    \\ \hline

\end{longtable}
\clearpage
\section{Feature Importance Graphs} \label{ac}
\FloatBarrier
 \begin{figure}[H]
\centering
\includegraphics[width=.65\columnwidth]{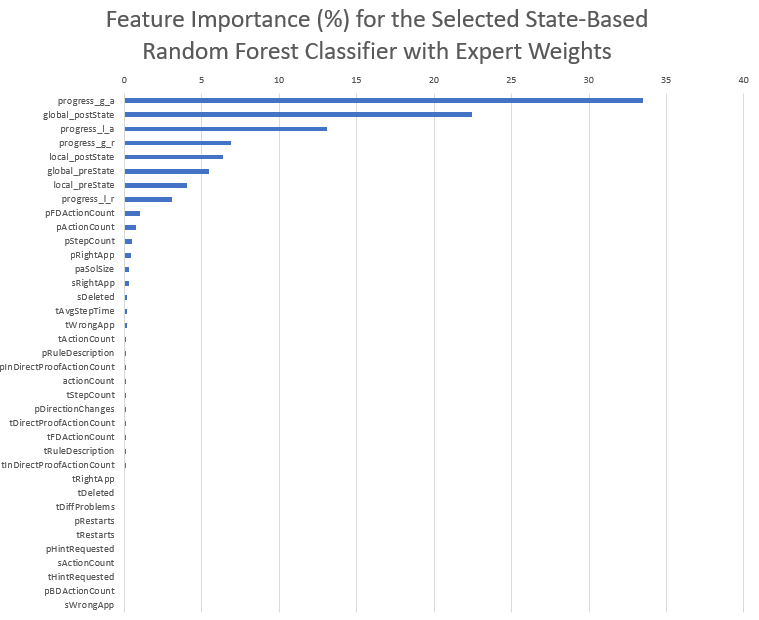}
\label{fig:sb_fe}
\end{figure}

 \begin{figure}[H]
\centering
\includegraphics[width=.70\columnwidth]{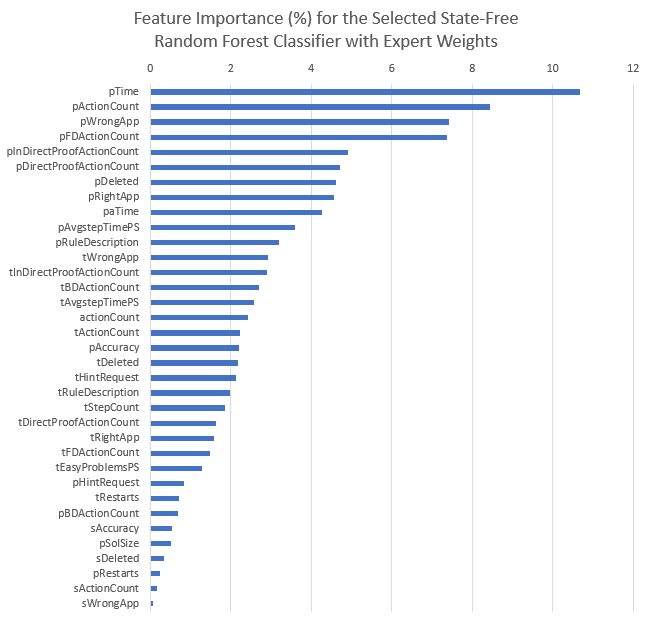}
\label{fig:sf_fe}
\end{figure}

\section{Comparing training steps in the two conditions on three aspects: prediction of HN/OK steps, observed HN/OK steps, and hint provision (on-demand or proactive)} \label{conf_mat}
\FloatBarrier
\begin{table}[H]
\centering
\begin{tabular}{|c|c|c|} 
\hline
Category             & \multicolumn{1}{l|}{Adaptive} & \multicolumn{1}{l|}{Control}  \\ 
\hline
pred-OK + noHints-OK & 4998                          & 2974                          \\ 
\hline
pred-OK + hinted-OK  & 106                           & 348                           \\ 
\hline
pred-HN + hinted-OK  & 1008                          & 232                           \\ 
\hline
pred-HN + noHints-OK & 0                             & 761                           \\ 
\hline
pred-HN + noHints-HN & 0                             & 1257                          \\ 
\hline
pred-HN + hinted-HN  & 812                           & 46                            \\ 
\hline
pred-OK + hinted-HN  & 51                            & 22                            \\ 
\hline
pred-OK + noHints-HN & 995                           & 335                           \\
\hline
\end{tabular}
\end{table}
\end{document}